# Double counting in LLM evidence fusion: exact posteriors separate rule compliance from posterior gain

Tengfei Shao[a], Asma Rezaei[b], Ali Reza Kamali[c]
[a] Global Education Center, Waseda University, 1-6-1 Nishi-Waseda, Shinjuku-ku, Tokyo 169-8050, Japan
[b] Faculty of Business and Law, School of Management, Anglia Ruskin University, Cambridge CB1 1PT, United Kingdom
[c] Cambridge Silicon Age Centre for Innovation, Barnwell Business Park, Cambridge CB5 8UZ, United Kingdom
Corresponding author: Tengfei Shao (tengfei.shao@toki.waseda.jp)

## Abstract

Language models fuse reports in retrieval, multi-agent consensus and agent memory, where treating shared-source reports as independent can double count evidence. A common remedy is an instruction to count each source once. Its value depends on the receiver's raw baseline. Pricing it needs an exact reference. We build the graded-dependence exact reference: binary worlds with single-report reliability 0.65 and four dependence levels, whose disclosed parameters give every world an exact posterior and every rule a target risk. The paired gain protocol scores raw and instructed answers of four configurations on 2,976 worlds. Agreement and gain separate in both directions. At $\delta = 0$, claude-opus-5-5 and gpt-6.1-sol agreed with the rule within 0.001 in 744 of 744, and the instruction raised their posterior mismatch by 0.044; gpt-6-luna agreed in 165 of 744 and gained 0.0140 (95% interval 0.0064 to 0.0219); claude-haiku-5-5's gain moved from −0.0295 ($\delta = 0$) to +0.0302 ($\delta = 1$) (descriptive). The prespecified first-child endpoint contrast, gain at $\delta = 1$ minus $\delta = 0$, was 0.0342 (0.0303 to 0.0382). In the preregistered tests, the raw source-weight contrast was positive (0.259, 95% interval 0.182 to 0.337; Holm rejected zero) and the grouping endpoint difference was not detected (0.0004, −0.0023 to 0.0031). This interval excludes differences larger than 0.0031 in magnitude and lies within the registered precision bound of ±0.0046. The reference prices the rule before any model call. The paired gain, estimable as a Brier difference on labelled deployment items, decides adoption per receiver and prompt condition.

## 1. Introduction

Double counting is a long-standing failure mode of information fusion. When reports share a source of error, combining them as if they were independent overstates the evidence, and correlated data are a recognised challenge across multisensor fusion [1]. When the dependence is known, fusion theory models it in the combination [2, 3, 4]. Track-to-track fusion, for instance, accounts for the common process noise between sensors [2]. When the dependence is unknown, fusion uses conservative idempotent operators that trade information for consistency. Covariance intersection keeps a fused estimate consistent when the cross-correlation is unknown [5], and the cautious rule combines belief functions whose bodies of evidence may overlap [6]. The conservative rule family of Taylor and Bishop [7] seeks closer agreement with the correlation-aware Bayesian result under unknown correlation while avoiding double counting. Large language models (LLMs) now perform this fusion step in retrieval-augmented generation, multi-agent consensus and agent memory, where several texts often descend from one source. A common remedy counts each source once, by keeping one representative per source family at retrieval [8] or by a prompt-level instruction [9]. The instruction hands an idempotent rule to a native receiver: a deployed LLM used as is, without task-specific training. Whether a pipeline should adopt the instruction is a decision about one receiver. The instruction's value depends on two things: the rule's target risk and the way the receiver fuses before the instruction. The check that is easiest to compute, agreement between the receiver's answers and the rule's output, measures numerical proximity to that output, whereas the decision needs the gain relative to the exact posterior. This paper measures both on the same worlds and shows that they separate in both directions: full agreement can come with a loss, and near-zero agreement with a gain.

When errors are normal with a common variance and correlation ρ, k dependent sources are worth $k/[1 + (k - 1)\rho]$ independent ones [3]. Verbatim copies add zero precision, noisy draws from a shared source add a positive and decreasing amount, and binary deduplication does not capture those graded increments [10]. Evaluations of LLM aggregators against exact or rational references mostly use repeated items with zero increment [11, 12, 13, 14], where counting once recovers the exact posterior. A noisy report from a shared source has a finite increment, so counting once discards some evidence, and an instruction can err by over-counting or over-discounting. Evaluations of LLM receivers lack an exact reference for graded same-source dependence. With one, a specific instruction delivered to a native receiver can be scored: how the receiver's distance from the exact posterior changes, and how that change relates to the rule's target risk and to the receiver's agreement with the rule.

We call the generator, together with its enumerated exact posteriors and rule target risks, the graded-dependence exact reference. Its worlds fix single-report reliability and vary the dependence between the reports of one source: a binary state is reported by two sources through three child reports each. The prompt discloses the generating parameters by design (Section 8.4), so every world has an exact posterior given the information shown. Any rule then has a target risk that a short program computes from the sampled world and the reference, without a model call. Like a calibration phantom, the reference supplies a known target. We call the paired raw-versus-instructed evaluation, with the risk expansion of Eq. (3), the paired gain protocol (Section 4.1). We instantiate count-once as a first-child rule (Section 3.3). Four closed-weight configurations answered 2,976 worlds under raw presentation (R), grouping by source (G), the count-once instruction (CO) and a dependence-aware note (DA). Before any model call, the reference fixes a forecast: a receiver whose raw answers already lie at b and who implements the rule faithfully pays the rule's target risk. The run shows whether each receiver meets the forecast, departs from it or reverses it, three cases that agreement counts cannot tell apart. The protocol applies when a count-once instruction is chosen before the receiver's baseline is measured.

Fig. 1 pairs the raw and CO answers of the same receiver on each world and scores them against two references computed without a model call: the first-child output and the exact posterior b. Agreement compares CO answers with the first-child output. Paired gain compares their posterior mismatch with that of the receiver's own raw baseline. At conditional independence ($\delta = 0$), where the three reports of a source share a source label and err independently, claude-opus-5-5 (Opus) and gpt-6.1-sol (Sol) agreed with the first-child output within 0.001 in 744 of 744 worlds each. The instruction raised their mismatch with the exact posterior by 0.0440 and 0.0441. At the same $\delta$, gpt-6-luna (Luna) agreed in 165 of 744 worlds and its paired gain was +0.0140 [0.0064, 0.0219]. For claude-haiku-5-5 (Haiku) the gain moved from −0.0295 at $\delta = 0$ to +0.0302 at $\delta = 1$ and changed sign between $\delta = 0.3$ and 0.6. At $\delta = 0$, where the rule's target risk is largest, ranking these receivers by agreement with the rule reverses their ranking by paired gain (descriptive). The same instruction is therefore worth a different amount to each receiver, and the amount depends jointly on the receiver's raw baseline and its response to the instruction. Section 6 reads each pattern through Eq. (3).

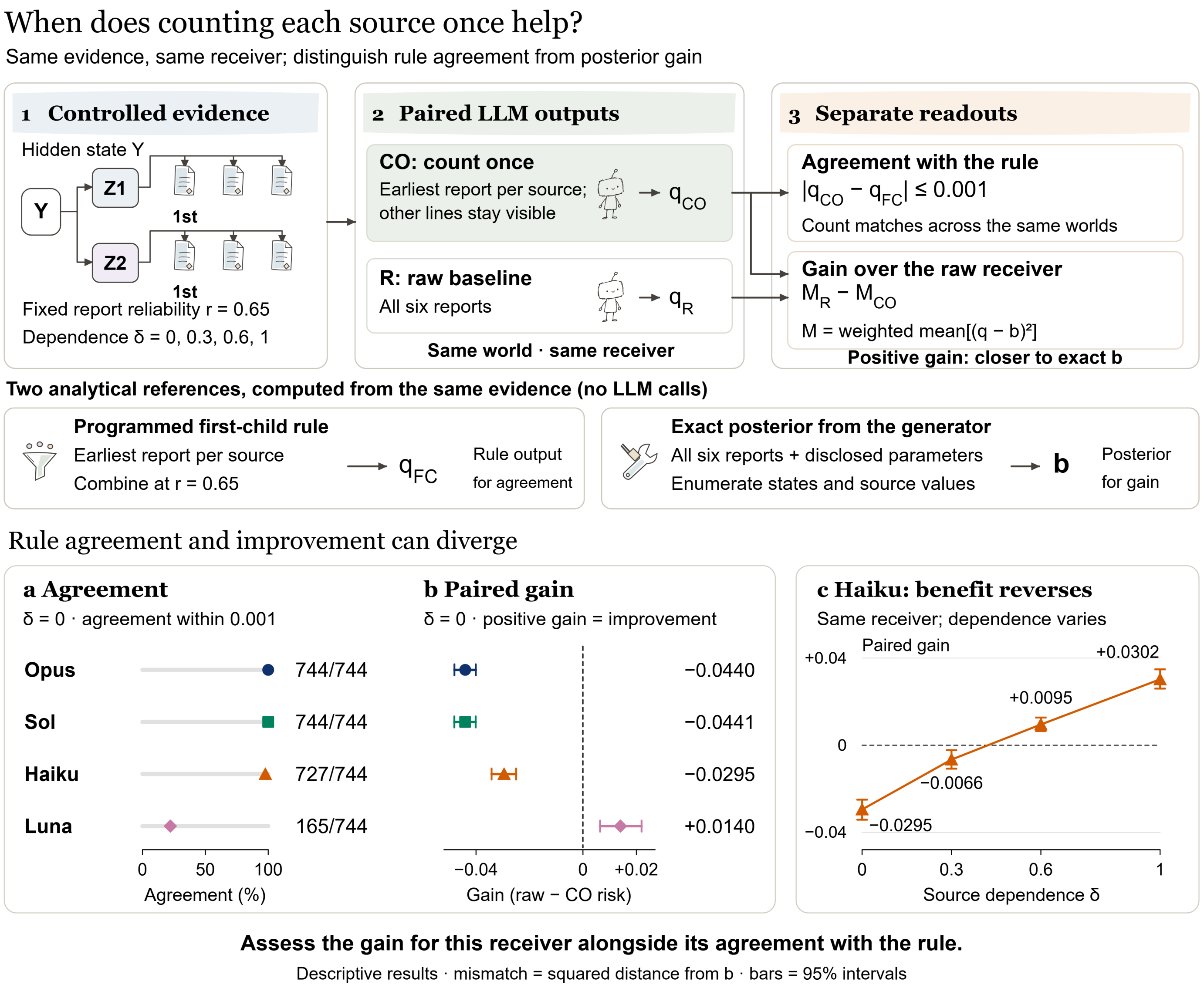

**Fig. 1. Rule agreement and paired gain separate in both directions.** Top: each world has a hidden state, two source values and three reports per source, with single-report reliability $r = 0.65$ and dependence $\delta = 0$, 0.3, 0.6 or 1. The same receiver answers each world in paired prompts, under raw presentation (R) and the count-once instruction (CO). All six reports remain visible, and CO asks the receiver to count each source's earliest original report once, as correct with probability $r$. $q_R$ and $q_{CO}$ are the receiver's reported probabilities. Two references are computed without a model call. The first, $q_{FC}$, is the output of the first-child rule, which combines each source's earliest report at $r = 0.65$. The second, $b$, is the exact posterior, obtained by enumerating hidden states and source values given all six visible reports and the disclosed generating parameters. Agreement means $|q_{CO} - q_{FC}| \leq 0.001$. Paired gain is $M_R - M_{CO}$, the fall in mismatch with $b$ relative to the receiver's own R answer, where $M$ is the natural-distribution weighted mean squared distance from $b$. Bottom: the four closed-weight configurations under CO. (a) Agreement at $\delta = 0$, counted in the cell-quota sample of 744 worlds per $\delta$. (b) Paired gain $\Delta_{CO} = E[(q_R - b)^2 - (q_{CO} - b)^2]$ at $\delta = 0$, with 95% stratified percentile intervals from 10,000 world-level resamples. Positive values mean smaller mismatch with the exact posterior than under raw presentation. Luna's compliance error $E[\eta^2]$ at $\delta = 0$ is 0.0341. (c) Haiku's paired gain at $\delta = 0$, 0.3, 0.6 and 1, with a line joining the measured levels (Fig. 4 gives the gradient for all four configurations). All quantities are descriptive.

The study makes three contributions to fusion under source dependence.

First, a finding about count-once fusion in LLM receivers: agreement with the rule and gain from it separate in both directions, so the value of the instruction is set by the receiver it is given to. The paired gain identifies the receivers for which it pays.

Second, the graded-dependence exact reference: a binary generator with fixed single-report reliability and finite same-source increments. Every report pattern has an exact posterior, and every candidate rule, from full count and first child to root majority, deduplication and an overlap-factor discount, has an exact target risk before any model call. The LLM runs answer whether this idempotent instruction pays for a given receiver.

Third, the paired gain protocol: the paired gain is the raw risk minus the CO risk, and the CO risk is the rule's target risk plus the receiver's squared deviation from the rule plus a cross term (Eq. (3)). Any gain or loss is therefore read from the receiver's raw risk together with these three terms. On labelled deployment items the Brier difference between the raw and instructed arms estimates the paired gain of Eq. (2) on those items without b (Section 8.3). The full closed-weight matrix cost USD 99.45, and its archived responses support a rerun of every analysis.

In the preregistered confirmatory family, the raw source-weight contrast was positive ($D_R = 0.259$ [0.182, 0.337]; Holm control rejected $D_R = 0$), and the grouping endpoint difference was not detected ($T_G = 0.0004$ [−0.0023, 0.0031]). This interval excludes endpoint differences larger than 0.0031 in magnitude and lies within the registered precision bound of ±0.0046.

## 2. Related work

### 2.1 Fusion of correlated and non-distinct sources

Forsling et al. [15] review a quarter century of covariance intersection and the cost of its conservatism while correlations stay unknown. IFNet [16] learns fusion weights from data when the correlation among sensor measurement noises is unknown. For probability judgements, consensus models allow stochastic dependence between sources [4], and Clemen and Winkler [3] and Bara [10] give graded results for Gaussian aggregators. Data integration infers the dependence between sources from the data and weights their votes accordingly [17], and truth discovery estimates source reliability [18, 19]. For belief functions, the cautious rule combines bodies of evidence that may overlap [6]. Idempotent conjunctive combination [20] and combination without the assumption of independence [21] extend the family. Fusion theory interpolates between independent and fully dependent evidence through the split covariance intersection filter [22], which separates an independent error component from one of unknown correlation, and the cautious-adaptive rule for partially non-distinct beliefs [23]. The disclosed generator, whose child reports are a shared source plus independent flips, gives that interpolation an exact target at every dependence level. In retrieval fusion, MoRE-RAG derives Bayesian weights assuming retriever scores are conditionally independent given relevance [24], and a design principle pools calibrated log-likelihood ratios to correct the count-scale drift of unnormalised weight sums [25].

For correlation that is non-zero and unknown, Taylor and Bishop [7] introduce a conservative family based on homogeneous functionals. Here the prompt discloses the dependence, so the exact posterior is the known-dependence optimum against which a receiver given a count-once instruction is scored. The first-child rule is idempotent, as covariance intersection [5] and the cautious rule [6] are. On a split source it keeps one report, where those operators pool all inputs. Any rule that counts an agreeing source as one independent report has a positive target risk at every δ below 1, where the exact weight w3 of an agreeing source (Section 3.1) exceeds one independent report. Such a rule is exact at δ = 1. First child and root majority are two examples.

Fusion gain and fusion harm can also be counted separately. Zhou and Lu [26] study decision-level fusion in vision–language recognition, where an auxiliary source can contest a recogniser's default prediction. Adopting every disagreeing challenger overwrites a correct answer in 18–46% of conflicts, and a Rescue–Hurt decomposition of keep-or-flip decisions makes harmful overwrites measurable. The present paper asks the same question, whether fusion helps or hurts, against an exact posterior. It measures both directions of a fusion instruction's effect on one receiver, and Section 8.2 reads both directions for each receiver.

### 2.2 LLM aggregation of repeated and dependent evidence

Evaluations of LLM aggregators with zero-increment repeats [11, 12, 13, 14] and of lineage in medical verification [9] report outcomes that depend on the model and the setting, with over-counting and over-crediting in [12] and [9]. A distributional account of evidence integration places the same dependence in the receiver: identical evidence can improve weaker models and harm stronger ones [27]. Krishna et al. [9] and Zhang et al. [28] show that recognising lineage or reliability cues does not ensure their use. Source statements can reduce the shift on distinct documents by 0.04–0.12 log-odds [11], which fits the observation here that an instruction can weaken evidence that is not a copy.

Table 1 places the contribution among the closest studies. The first two rows are the peer-reviewed anchors: graded dependence for Gaussian aggregators (Clemen and Winkler [3]) and the operators for unknown or partially non-distinct dependence (covariance intersection [5], the cautious rule [6] and the cautious-adaptive rule [23]). The graded-dependence exact reference adds an exact posterior for finite, graded same-source increments and a native receiver measured against its own raw baseline. It puts a rule's target risk, the receiver's output deviation and the paired gain on one scale. The comparison with the state of the art has two layers. Table 1 compares what each study can evaluate. Sections 3.3 and 3.4 price the existing remedies (full count, first child, root majority, deduplication and the overlap-factor discount adapted from [29] to this source partition) on one exact scale under the same evidence model.

**Table 1.** The evaluation gap relative to the closest studies

| Study | Dependent increment | Reference | Intervention locus | What the present study adds |
|---|---|---|---|---|
| Clemen and Winkler [3] | Graded; correlation ρ among normal errors | Exact precision of the Gaussian model | Closed-form aggregator | A binary generator with an exact posterior per cell; native LLM receivers |
| Covariance intersection [5]; cautious rule [6]; cautious-adaptive rule [23] | Unknown or partially non-distinct | Covariance consistency [5]; idempotent combination of overlapping evidence [6]; adaptive combination of partially non-distinct evidence [23]; no exact posterior per pattern | Fusion operator | An exact posterior per report pattern and an exact target risk for each rule |
| Epistemic Sybil Resistance [10] | Graded; exact in an equicorrelated Gaussian model | Exact precision of the Gaussian model | Closed-form aggregators over agent estimates | A binary generator with an exact posterior per cell; native LLM receivers |
| Provenance-Aware Belief Updating in Human-AI Feedback Loops [30] | Relayed reports with supplied lineage | Bayesian supervision | Models trained towards the target | Native receivers that are not trained; a prompt-level instruction scored against an exact posterior, with a paired raw baseline |
| Consensus Is Not Evidence [29] | Partial document overlap across retrieval lineages | Task answers (accuracy, negative log-likelihood) | Aggregator outside the model | A prompt-level instruction scored against an exact posterior, with a paired raw baseline |
| Copies or Sources? [11] | Zero for copies; a gradient to model-written summaries without an exact posterior | Exact Bayes posterior over conditionally independent sensors | LLM aggregator; source statements and a reference rule in prompts | Finite noisy same-source increments at four dependence levels, each with an exact posterior |
| Record Grouping [31] | Textual partial copies with complementary content in a group | Unit-weight reference | LLM; external representative and prompt rules | A numerical posterior reference; first child as one representative rule with a computable risk |
| What Counts as Evidence? [28] | Binary lineage contrast | Matched choice-shift contrasts; cue identification | LLM; restatement, ledger and weighting rule | Size of the response relative to an exact target; agreement and gain reported separately |
| This study | Finite and graded, exact for every cell | Exact posterior from the information the prompt shows | Native receiver; grouping, first-child rule and note in the prompt | Rule target risk, output deviation and paired gain on one scale |

Notes: LLM, large language model. Entries summarise the versions cited.

Exact Bayes targets also underlie work that trains models on probabilistic reasoning [32], and probability tasks have been evaluated through estimation, sampling and probability calculations [33]. Human studies of correlation neglect [34], persuasion bias [35] and the illusion of consensus [36] identify the behaviours that motivate count-once instructions, as do LLM studies of source preference under repetition [37] and of knowledge conflicts [38]. Memory, retrieval and message-passing designs weight shared-source reports at a fusion step that the protocol can measure: they keep one representative per provenance family [8], diversify retrieval [39], admit items to shared memory [40] or limit repeated routing [41]. Information Fusion has published an empirical analysis of uncertainty measures for fusing crisp and fuzzy bodies of evidence [42] and a study of in-context learning as the fusion of two information sources in a controlled setting [43]. The present design adds an exact posterior per world.

## 3. The graded-dependence exact reference

### 3.1 The generator and the dependence gradient

A world has a hidden binary state Y with prior $\pi$ drawn from {0.25, 0.50, 0.75}. Two sources Z1 and Z2 are independent given Y, and each equals Y with probability a. Each source emits three child reports $X_j = Z_g \oplus E_j$, with flips $E_j$ that are Bernoulli(e) and independent given the sources (Fig. 2a). The 3+3 partition, source labels, senders and arrival positions are randomised, and Y enters the analysis through scoring alone. The single-report reliability is $r = P(X_j = Y) = a(1 - e) + (1 - a)e$, and the dependence $\delta$ is the correlation between the correctness of two child reports of one source. Given r and $\delta$, the parameters a and e are determined (Supplementary Section S1.1), so $\delta$ governs how much a source adds beyond its first report while r stays fixed.

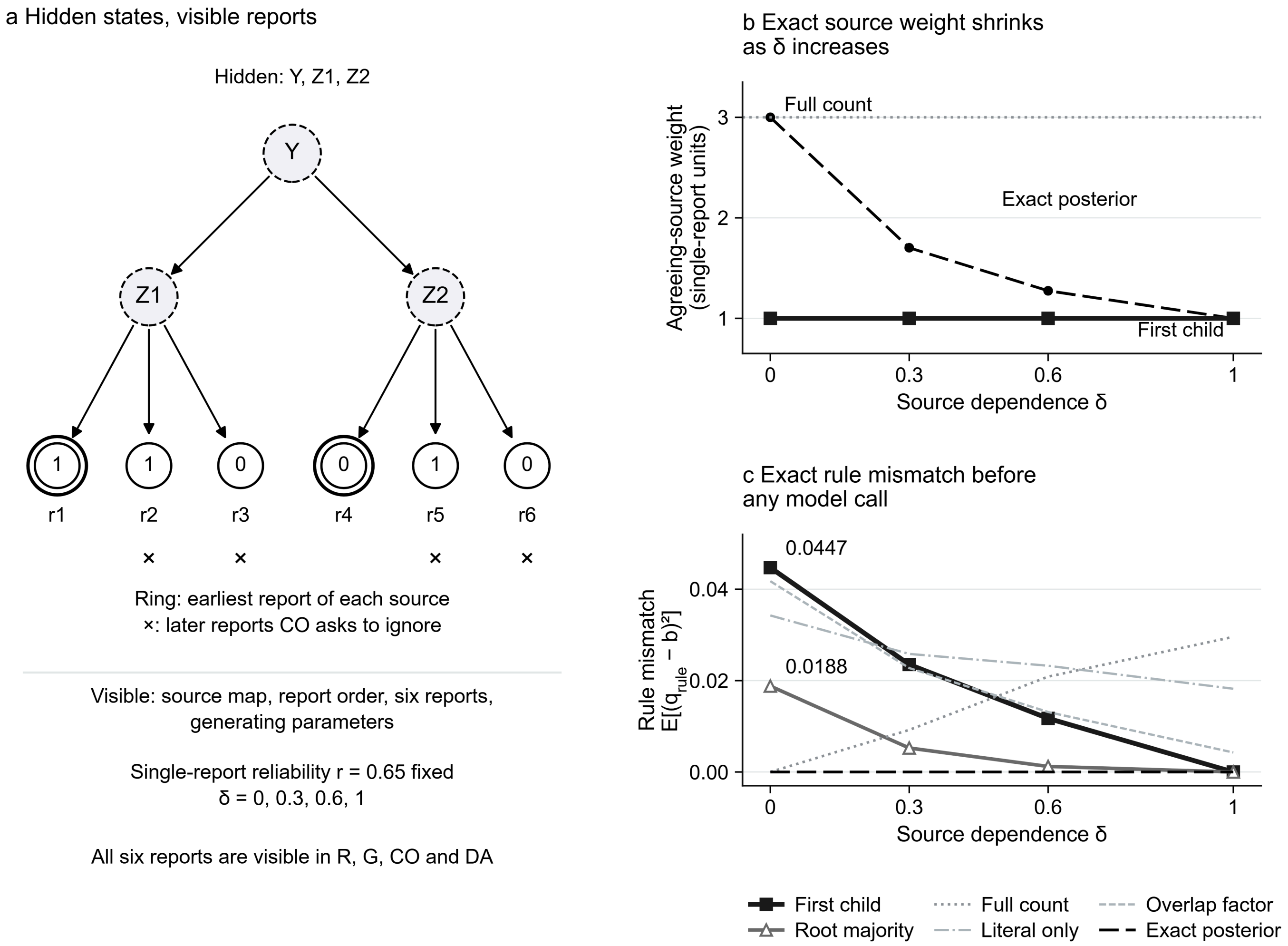

**Fig. 2. Known source dependence gives an exact target for each rule.** (a) The generator, illustrative reports and the first-child selection. A hidden state Y is reported through two sources, Z1 and Z2, each with three child reports drawn as white circles with a black 0 or 1. A double ring marks the earliest report of each source, which the first-child rule counts. A cross marks a later report that stays visible but that the count-once instruction (CO) tells the receiver to ignore. Reliability r = 0.65 is fixed and dependence δ takes the values 0, 0.3, 0.6 and 1. The receiver sees the six reports under raw presentation (R), grouping by source (G), CO or the dependence-aware note (DA). The exact posterior b is enumerated over the 64 report patterns. At δ = 1 no source can split 2:1, so patterns with a split source have likelihood 0 and are excluded. (b) Weight w3 of an agreeing source (3:0), in units of one independent report, against δ: exact 3.000, 1.702, 1.274, 1.000; first child 1; full count 3. First child and root majority share w3 = 1 and differ in the signed log-odds weight w2 of a 2:1 split source, averaged over random arrival orders (1/3 and 1; w2 is undefined at δ = 1). (c) Exact population mismatch of all program rules against δ, with first child (0.0447, 0.0235, 0.0117, 0) and root majority (0.0188, 0.0052, 0.0012, 0) emphasised. The realised-world target risk used in Eq. (3) is 0.0440754, 0.0229553, 0.0122546 and 0. The legend entry Overlap factor is the overlap-factor discount rule of Supplementary Section S1.2, built on the overlap factor of RIDE-CAL [29]. All curves are exact and computed without a model call.

On the log-odds scale, a conditionally independent report contributes ±logit r, a verbatim copy contributes zero, and successive agreeing reports of one source contribute amounts in between that shrink as δ grows. At r = 0.65, the second and third agreeing child reports add 1.000 and 1.000 reports at δ = 0, 0.527 and 0.176 at δ = 0.3, 0.243 and 0.031 at δ = 0.6, and nothing at δ = 1, so an agreeing source has weight w3 = 3, 1.702, 1.274 and 1 in units of one independent report (Fig. 2b; Supplementary Section S1.1). The Gaussian result [3, 10] gives the corresponding continuous discount. Here every increment is exact for binary reports. With ρ set to δ, the Gaussian factor 3/(1

+ 2δ) agrees with the binary weight at both endpoints (3 and 1) and gives 1.875 and 1.364 between them, against 1.702 and 1.274 here.

At $\delta = 0$, a = 1 and e = 0.35, so each source equals Y and the reports are independent given Y. The source map labels three reports as sharing a source, so a receiver that discounts by label alone loses information. At $\delta = 1$, e = 0 and all reports of a source agree, so no source can split 2:1. For every δ below 1, a 2:1 split source is worth exactly one report under the exact posterior, so dependence acts on agreeing sources. With r fixed across δ, the Bayes risk is 0.1331, 0.1543, 0.1661 and 0.1778 at δ = 0, 0.3, 0.6 and 1 (Supplementary Section S1.2), and Section 6.5 standardises on common difficulty support.

### 3.2 The exact posterior and the mismatch score

The exact posterior is $b = \pi L1/[\pi L1 + (1 - \pi)L0]$, where $L_y$ is the likelihood of the visible report pattern under state y with each source summed out (Supplementary Section S1.1). All 64 patterns are enumerated, and b is defined for the patterns that have positive likelihood. The prompt states π, the generating mechanism including the conditional independence structure, a and e rounded to four decimals, r and δ, every report's origin ID, and which reports share a source. The state, a source's value, the posterior, the arm and the world ID never appear. Because the receiver is told the mechanism, b is the posterior under its own information.

The score is the mismatch $(q - b)^2$, where q is the parsed reported probability. Under the generator, P(Y = 1 given the visible prompt x) equals b, and q is conditionally independent of Y given x, so

$$E[(q - Y)^2 \mid x, q] = b(1 - b) + (q - b)^2 \qquad (1)$$

Mismatch is the expected Brier score in excess of the Bayes risk $b(1 - b)$. Because the Brier score is strictly proper [44], this excess is the natural scale for departures from b. Logit quantities use the clipped logit with $\varepsilon = 0.001$, and $\varepsilon = 10^{-6}$ is the sensitivity scale (Supplementary Section S2.3).

### 3.3 Presentation arms and the first-child rule

Arm R lists the two sources and the six reports in arrival order. Arm G shows the same six reports grouped under source headings, so G differs from R in order and organisation. Arm CO adds one sentence, the count-once instruction: "for each source, use only its earliest original report and count it once as a report that is correct with probability r; ignore that source's other lines", where earliest means arrival order. The first-child rule is the program that applies the same selection: it combines each source's earliest report at r, and $q_{FC}$ denotes its output, the first-child output. All six reports remain on the page. The instruction sets reports aside, and the layout keeps them. Arm DA adds a note, without numbers, that a copy or rewording adds no new information and that the contribution of different original reports from one source is set by the generating process, so the receiver should avoid both over- and under-counting. Five further program rules are computed without a model call on the same visible information: full count, root majority, literal-only deduplication, origin-aware deduplication (which equals b) and an overlap-factor discount built on the overlap factor of RIDE-CAL [29] (Supplementary Section S1.2).

### 3.4 What a count-once instruction costs before any model call

Averaging over every order in which a source's reports can arrive, the first-child rule has population mismatch 0.0447, 0.0235, 0.0117 and 0 at $\delta = 0, 0.3, 0.6$ and 1 (Fig. 2c). It is idempotent (Section 2.1): three identical reports retain the evidence of one report, and its agreeing-source weight is $w3 = 1$ in units of one independent report. It is exact at full dependence and underweights agreeing sources below it. Its mismatch is larger than that of full count at $\delta = 0$ and 0.3 and smaller at $\delta = 0.6$ and 1 (full count: 0, 0.0092, 0.0209 and 0.0296). Root majority also keeps one value per source and has the same agreeing-source weight ($w3 = 1$). It weights a 2:1 split source differently: averaged over random arrival orders, the signed log-odds weight is $w2 = 1/3$ for first child and 1 for root majority. The cost therefore belongs to the selection and weighting rule. Root majority has mismatch 0.0188, 0.0052, 0.0012 and 0: smaller than that of first child at each $\delta$ below 1, less than half of it at $\delta = 0$, and equal to it at $\delta = 1$. The generator prices any candidate rule exactly and ranks the two rules before any model call.

On the 2,976 sampled worlds, the earliest reports are those realised, and the first-child target risk, which follows from the sampled worlds and the reference alone, is 0.0440754, 0.0229553, 0.0122546 and 0. The population curve prices the rule before any sample is drawn, and Section 4.1 uses the realised value.

## 4. The paired gain protocol

### 4.1 Agreement, mismatch and paired gain on one set of worlds

Three quantities are reported on the same worlds for the count-once instruction: the raw posterior mismatch, the agreement of CO answers with the rule's output, and the paired gain. A CO answer agrees with the rule when it lies within 0.001 of $q_{FC}$ (Supplementary Section S2.3). In worlds where the rule's output equals the stated prior, a receiver that repeats the prior agrees with the rule. Agreement is therefore also tabulated over the worlds where the two differ (415, 406, 404 and 414 of 744 at the four levels). For an arm X the paired gain is

$$\Delta_X(\delta) = E[(q_R - b)^2 - (q_X - b)^2] \quad (2)$$

where $q_R$ and $q_X$ are one receiver's answers to the same world under raw presentation and arm X, so a positive value means arm X gives that receiver smaller mismatch than raw presentation. The risk of an arm is its mismatch averaged over the natural distribution of worlds defined below.

Worlds are sampled by cell quotas. A cell combines the prior with the unordered pair of source counts of reports saying 1 (Supplementary Section S1.3). The sample contains 744 worlds per $\delta$ and 2,976 overall. Agreement counts describe this quota sample. Risks and gains average over the natural distribution. Here L denotes the wording frame. Under L0 every child report uses one shared sentence frame. Under L1 each arrival position uses a different one of six fixed meaning-preserving frames. Each $\delta \times \pi$ combination contains equal numbers of worlds under the two frames (Supplementary Section S1.3). Within each $\delta \times \pi \times L$ layer, exact cell probabilities weight the cell means. The six $\pi \times L$ layers at each $\delta$ (three priors, two frames) receive equal weight. Counts and gains retain their respective denominators.

To relate the three quantities, write $\eta = q_{CO} - q_{FC}$ for the deviation of a CO answer from the rule's output. Expanding the square on the same worlds and with the same weights gives

$$E[(q_{CO} - b)^2] = E[(q_{FC} - b)^2] + E[\eta^2] + 2E[\eta\,(q_{FC} - b)] \quad (3)$$

The CO risk is thus the rule's target risk plus the squared deviation plus the cross term $2E[\eta(q_{FC} - b)]$, which is negative when deviations point towards b and positive when they point away. The squared deviation $E[\eta^2]$ is the compliance error, a prespecified secondary quantity. With the agreement count, it measures rule compliance. The paired gain $\Delta_{CO}$ is the raw risk minus the CO risk, so $\Delta_{CO}$ = (raw risk − target risk) − (CO risk − target risk), an exact split used as a reading tool. The first term is the reference gain: the gain of a receiver that implements the rule faithfully, determined by the raw arm and the reference. It equals $\Delta_{CO}$ when $\eta = 0$. The second term is the implementation term: the squared deviation plus the cross term, measured by the instructed arm. Eq. (3) uses the realised-world target risk of Section 3.4 (Supplementary Section S2.3).

### 4.2 Configurations and panels

Table 2 lists the six fixed configurations. The four closed-weight configurations share the registered template P0-draft-2, the same 2,976 main-panel worlds and all four arms, and form the panel for the confirmatory family. The two open-weight configurations answered a revised template (P0-draft-3) and extend the panel to two further model families. Each configuration is a deployed unit of a model and its vendor reasoning setting, and comparisons are between configurations as deployed (Supplementary Section S6.4). For commercial endpoints whose behaviour changes over time [45], the archived 57,600 closed-weight responses, the generator, the exact references and the analysis code support a rerun of every analysis without a model call. No sampling parameters are sent. Each request is the full rendered prompt as a user message in a fresh context, with no study-supplied system message, tools or memory.

**Table 2.** Study design, configurations and outcomes across the three formal panels

| Configuration | Reasoning setting | Template and output cap | Valid (of 14,400) | Failures | Settled cost (USD) | Analysis role |
|---|---|---|---|---|---|---|
| **Closed-weight configurations** | | | | | | |
| claude-opus-5-5 (Opus)[a] | Adaptive thinking (always on); effort low | P0-draft-2; 4,096 tokens including reasoning | 14,400 | 0 | 80.06 | Confirmatory panel |
| gpt-6.1-sol (Sol)[a] | Reasoning effort low | P0-draft-2; 4,096 tokens including reasoning | 14,400 | 0 | 17.70 | Confirmatory panel |
| claude-haiku-5-5 (Haiku)[a] | Thinking disabled; effort low | P0-draft-2; 4,096 tokens including reasoning | 14,400 | 0 | 1.26 | Confirmatory panel |
| gpt-6-luna (Luna)[a] | Reasoning effort none | P0-draft-2; 4,096 tokens including reasoning | 14,389 | 11[b] | 0.43 | Confirmatory panel |
| **Open-weight configurations** | | | | | | |
| Qwen2.5-32B-Instruct-FP8-dynamic (Qwen)[c] | No thinking mode | P0-draft-3; 256 visible tokens | 14,400 | 0 | n/a | Replication |
| gemma-3-27b-it-FP8-dynamic (Gemma)[c] | No thinking mode | P0-draft-3; 256 visible tokens | 14,400 | 0 | n/a | Replication |

Notes: Each configuration had 14,400 scheduled items: main panel 2,976 worlds × 4 arms = 11,904; r = 0.70 panel 768 worlds × 2 arms = 1,536; wording-control panel 240 parent worlds × 2 variants × 2 arms = 960. Across six configurations this is 86,400 final items (closed 57,600, open 28,800). Validity counts are final outcomes after the registered replay rule. Settled cost is at Batch prices. The closed-weight total is USD 99.45. Prices and snapshot retirement dates are in Supplementary Section S6.4. Opus and Haiku are Anthropic models and Sol and Luna are OpenAI models. The open-weight repositories carry the prefix RedHatAI/. Thinking disabled refers to the API parameter. Visible text before the JSON answer was permitted. Configuration-level CO and DA trajectories and the open-weight replications are descriptive. [a] Message Batches API (Claude) or Batch API, /v1/responses (GPT). [b] Nine refusals (deviation D1: input-side content-filter rejections treated as refusals) and two truncations. Failed answers are filled with the prior π and flagged. At δ = 0 there are no R or CO failures in any closed configuration (744 valid per arm). [c] vLLM 0.32.0 on a GCP H100 (deviation D2). The open-weight template adds the answer sentence "Do not show your reasoning or write any other text." (deviation D3). Open-weight runs used GCP H100 machine time, with no API charge. n/a: not applicable.

## 4.3 Registered status and inference

Each reported quantity carries one stated status: confirmatory, prespecified secondary, descriptive or exploratory. The confirmatory family has two two-sided tests on the equal-weight mean of the four closed-weight configurations, adjusted by Holm's procedure [46] at a nominal family-wise level of 0.05, each with a positive expected direction stated in advance. Inference is over worlds for a fixed panel of four configurations. Cell-mean responses are summarised by a fixed common projection,

$$\operatorname{logit} q \approx \alpha \operatorname{logit} \pi + \operatorname{logit} r\,(w_3 U + w_2 S) \tag{4}$$

where U counts sources whose three reports all say 1 minus sources whose three reports all say 0, and S counts sources split 2:1 towards 1 minus sources split 1:2. The coefficients w3 and w2 give the weights of an agreeing source and of a split source, in units of one independent report. The exact posterior satisfies this form exactly. For other receivers, w3 and w2 are coefficients of the fixed projection (Supplementary Section S2.4).

The first confirmatory quantity is the source-weight contrast under raw presentation,

$$D = w_3(0.3)\, w_2(0.6) - w_3(0.6)\, w_2(0.3) \tag{5}$$

with $D_R$ its four-configuration mean in arm R. It is zero for any receiver that ignores δ and for one whose response to δ is a common rescaling of δ-independent weights. The exact posterior gives D = 0.428. The second is the endpoint difference of the grouping effect across the dependence gradient,

$$T_G = \Delta_G(1) - \Delta_G(0) \quad (6)$$

averaged over the four configurations (Supplementary Section S2.4).

The first-child endpoint contrast $T_{CO} = \Delta_{CO}(1) - \Delta_{CO}(0)$ and the saturation diagnostic $D_{diag}$ are prespecified secondary quantities. The descriptive results are the CO trajectories and their intervals from the formal analysis driver, the expansion of Section 4.1, and the per-δ agreement counts (post hoc descriptive tallies of the final answer logs). They include counts over the worlds where the rule's output differs from the prior (Supplementary Section S4.1). Result branches and wording rules are fixed in Supplementary Section S2.5.

Intervals and tests use a world bootstrap stratified by δ × π × L × cell, with B = 10,000 resamples, rescaled percentile intervals, and all arms and configurations of a world resampled together (Supplementary Section S2.5). Coverage calibration and the calibration-support screen concern the two confirmatory four-configuration quantities (Supplementary Section S3.3). Per-configuration, secondary and descriptive intervals are 95% percentile intervals, unadjusted for multiplicity, from the same resamples. Deviation sensitivity reports conditional envelopes for how far rare deviations could move the true values at budgets of K = 1 and K = 4 expected deviating worlds per configuration and arm (Supplementary Section S3.4).

**4.4 Registration, data status and deviations**

The study was preregistered on OSF on 28 October 2026 (https://osf.io/dzgj2/; the registration is under an approved embargo), and formal requests began the same day after the registration was complete. The analysis driver first ran on formal answers on 31 October 2026, after the closed-weight matrix and the open-weight replication were finished. The 2,976 formal worlds are disjoint from the development pilot, whose data are kept apart from every confirmatory and secondary estimate (Supplementary Section S6.1).

Supplementary Section S6 documents eight post-registration deviations from the registered rules (D1–D8) and six corrections to the registered code (C1–C6) that preserve the registered analysis definitions. D1–D3 are described in the notes to Table 2, D6 in the notes to Table 3, D8 in Section 6.6, and D4, D5 and D7 in Section 7.3.

## 5. Raw baselines and the registered family

The registered family tests two quantities for the four-configuration panel (Table 3). The raw source-weight contrast was positive: $D_R = 0.25941$, with a 95% interval of [0.1824, 0.3369]. Holm control rejected $D_R = 0$, which selects branch A+, the registered branch for a rejected, positive $D_R$. All four ν gates pass in arm R: at both middle levels, the split-source weight w2 and the total weight w3 + w2 are bounded away from zero. The grouping endpoint difference was not detected: $T_G$ = 0.00037, with a 95% interval of [−0.0023, 0.0031] and p = 0.7958. This interval excludes endpoint differences larger than 0.0031 in magnitude and lies within the registered precision bound of ±0.0046, a quarter of the endpoint difference in the planning scenario in which grouping tracks dependence. Under branch A+ the registered reading is the sign of $D_R$. The ratio change $\Lambda_R$ is the four-configuration mean of $\nu(0.3) - \nu(0.6)$, with $\nu = w3/(w3 + w2)$. Its descriptive estimate is 0.0274 [−0.0087, 0.0626]. The saturation diagnostic $D_{diag}$, 0.177 [0.081, 0.275], has an interval that excludes 0, the value of every receiver that acts on the reports through their count N alone (Supplementary Section S2.4).

**Table 3.** Registered family, prespecified secondary quantities, and scale and completeness checks

| Quantity or check | Result | 95% interval | p | Status |
|---|---|---|---|---|
| **Confirmatory family (Holm, two-sided, nominal α = 0.05)** | | | | |
| $D_R$, raw-arm source-weight contrast | 0.25941 | 0.1824 to 0.3369 (97.5%: 0.1704 to 0.3472) | 0/10,000 resamples cross zero | Rejected, positive (branch A+) |
| $T_G$, grouping endpoint difference | 0.00037 | −0.0023 to 0.0031 (97.5%: −0.0027 to 0.0035) | p = 0.7958 | Not detected; excludes differences larger than 0.0031 in magnitude; within ±0.0046 |
| **Prespecified secondary and descriptive** | | | | |
| $T_{CO}$, first-child endpoint contrast | 0.03419 | 0.0303 to 0.0382 | 0/10,000 resamples cross zero | Prespecified secondary; Holm flag within the secondary family |
| $D_{diag}$, saturation diagnostic | 0.17735 | 0.0813 to 0.2749 | p = 0.0002 | Prespecified secondary diagnostic |
| $\Lambda_R$, ratio change | 0.02742 | −0.0087 to 0.0626 | p = 0.1332 | Descriptive |
| **Scale and completeness checks** | | | | |
| $\varepsilon = 10^{-6}$ | $D_R$ 0.28738; $T_G$ 0.00037 | $D_R$ 0.1820 to 0.3943; $T_G$ −0.0023 to 0.0031 | $T_G$ p = 0.7958 | Both Holm decisions and branch unchanged |
| Complete cases (2,969 worlds) | $D_R$ 0.25761; $T_G$ 0.00035 | $D_R$ 0.1808 to 0.3369; $T_G$ −0.0023 to 0.0030 | $T_G$ p = 0.7984 | Both Holm decisions and branch unchanged |

Notes: N = 2,976 main-panel worlds, four closed-weight configurations weighted equally, B = 10,000 stratified world-level resamples, ε = 0.001 unless stated. Intervals are percentile intervals, and p is the empirical p value of the registered counting rule. Where no resample crossed zero the entry reads 0/10,000 resamples cross zero, and the smallest positive p of the rule is 0.0002. The 97.5% intervals target nominal simultaneous 95% coverage for the confirmatory pair by Bonferroni adjustment. Coverage calibration, the calibration-support screen, scenario X04 (deviation D6) and the deviation envelopes are reported in Supplementary Sections S3.3–S3.4 (Tables S3j and S3k).

The four-configuration mean combines three raw response patterns (Fig. 3), the baselines for Section 6. Opus and Sol have D = 0.4306 [0.4283, 0.4352] and 0.4310 [0.4279, 0.4372] against the exact 0.4280, w3 = 1.700 and 1.702 at δ = 0.3, and 1.274 and 1.275 at δ = 0.6, and projection residuals of at most 0.008 logit. Haiku weights agreeing sources above the exact value at every level (w3 = 3.908, 2.640, 2.177 and 2.914 at δ = 0, 0.3, 0.6 and 1, against 3, 1.702, 1.274 and 1), with D = 0.219 [−0.068, 0.508]. Luna leans towards the prior (regression coefficient β = 0.283 on the exact log-likelihood ratio, w3 of 0.672 and 0.782 at the middle levels) with D = −0.043 [−0.156, 0.069]. The D intervals of Haiku and Luna both include 0 (Tables S3a–S3e). At δ = 1 no source splits, so w2 is undefined and w3 comes from a separate projection, compared with its exact value at that level alone (Supplementary Section S2.4). Given Haiku's and Luna's weighted projection residuals at the middle levels (0.278 and 0.342 for Haiku, 0.306 and 0.487 for Luna), their w3 and w2 are read as fixed projection coefficients. A positive $D_R$ therefore comes mainly from Opus and Sol. The four raw baselines differ in kind, so Section 6 scores the instruction against each receiver's own raw baseline.

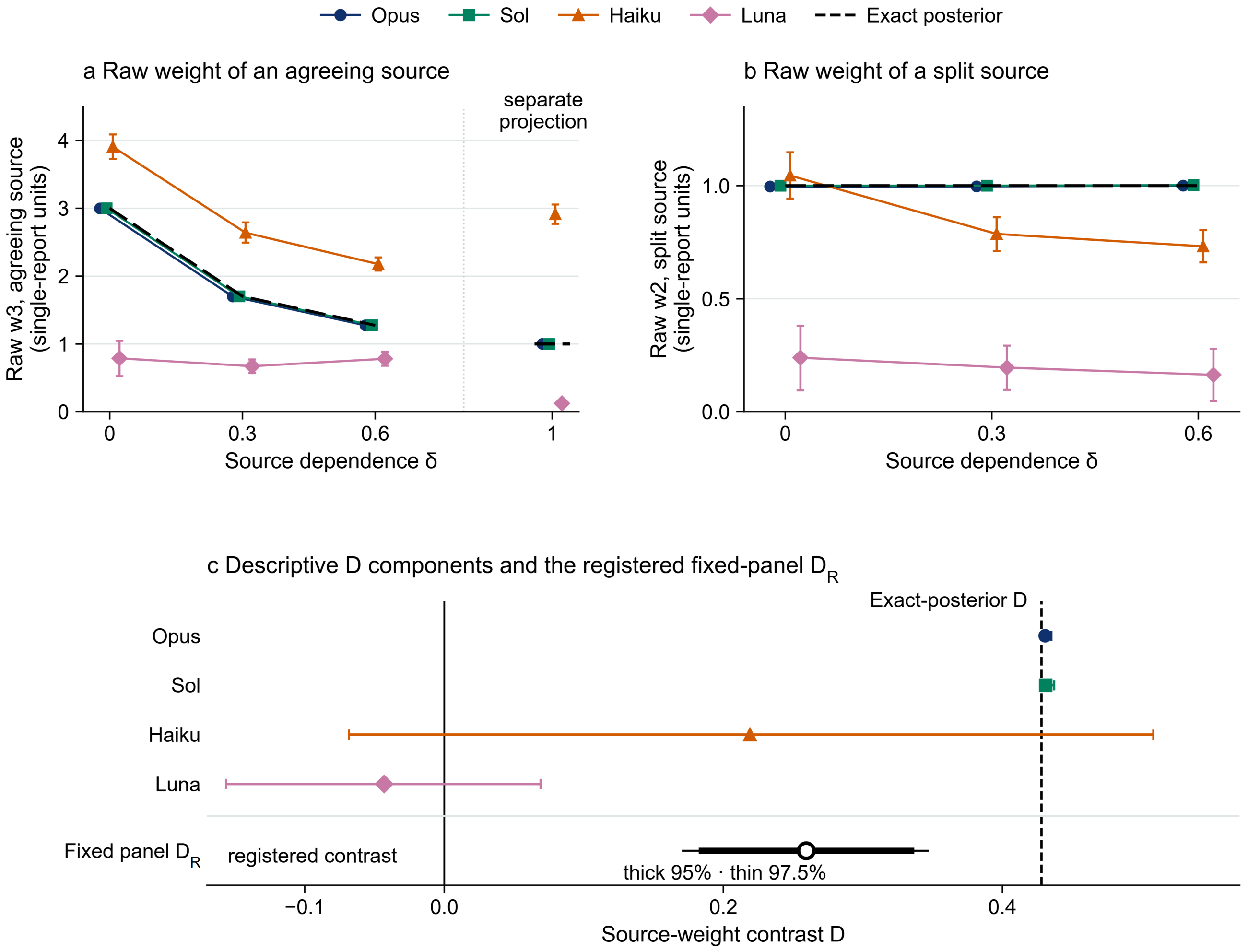

**Fig. 3. Raw receivers assign different weights along the dependence gradient.** (a) Projected source weight w3 of an agreeing source (3:0) under raw presentation (arm R) against δ for the four closed-weight configurations, with the exact reference (dashed line). The δ = 1 point comes from a separate projection and is not joined to the other levels. (b) Projected weight w2 of a split source (2:1) at δ = 0, 0.3 and 0.6. The weight w2 is undefined at δ = 1, where no source splits. (c) Source-weight contrast D (Eq. (5)) against the exact value 0.4280: Opus 0.4306 [0.4283, 0.4352], Sol 0.4310 [0.4279, 0.4372], Haiku 0.219 [−0.068, 0.508], Luna −0.043 [−0.156, 0.069]. Four-configuration mean $D_R$ = 0.259 [0.182, 0.337] (thin line: 97.5% interval [0.170, 0.347]). $D_R$ and $T_G$ form the preregistered confirmatory family ($T_G$ is in Table 3). The per-configuration values are descriptive. Other intervals are 95% stratified percentile intervals from 10,000 world-level resamples. N = 2,976 worlds, ε = 0.001. The panels show the raw baselines against which Fig. 1 scores the instruction.

$T_G$ concerns a layout change: grouping keeps all six reports and regroups them under source headings, so the contrast measures how the effect of that layout differs between the endpoints. The four-configuration mean paired gain of grouping was $\Delta_G$ = 0.0013, 0.0012, 0.0009 and 0.0017 at δ = 0, 0.3, 0.6 and 1, with an interval excluding 0 at δ = 1 ([0.0003, 0.0031]). At δ = 0.3 the mean combines opposite movements: grouping raises Haiku's mismatch ($\Delta_G$ = −0.0044 [−0.0081, −0.0004]) and lowers Luna's (+0.0091 [0.0036, 0.0145]). Under G and DA, Opus and Sol stay within 0.001 of the exact posterior in at least 2,970 of 2,976 worlds. The note's mean gains are $\Delta_{DA}$ = 0.0001, 0.0011, 0.0017 and 0.0015, with an interval excluding 0 at δ = 0.6 ([0.0002, 0.0032]). The ratio changes $\Lambda_G - \Lambda_R$ and $\Lambda_{DA} - \Lambda_R$ have intervals [−0.0270, 0.0571] and [−0.0513, 0.0378], and the secondary contrasts $D_G$ = 0.276 [0.194, 0.358] and $D_{DA}$ = 0.241 [0.167, 0.316] carry a Holm flag within the secondary family (Table S4g).

Both confirmatory quantities pass the registered calibration-support screen on the formal data, and Table S3h details the checks of Table 3. At the screen composition of the formal run, scenario X04 gives 95% coverage of 0.9516 for $D_R$ and 0.9493 for $T_G$ (Table S3j). The scenario was built after the formal results were seen (deviation D6). The conditional deviation envelopes (Table S3k) quantify stability at each deviation budget: $D_R$ stays positive in three of its four envelopes, including both rule-switch envelopes. The widened $T_G$ interval stays within the registered precision bound of ±0.0046 in one of the four envelopes (rule switches, K = 1). Closeness to b is reported per configuration as counts in Supplementary Section S3.5.

## 6. Instruction gain and rule agreement separate in both directions

Fig. 1 shows the three patterns for the four closed-weight configurations under identical prompts. Fig. 4 traces their paired gain across all four dependence levels against the rule's target risk. Table S4a gives agreement and paired gain at every dependence level. Gains carry the 95% stratified bootstrap intervals of Section 4.3. Section 6.4 applies Eq. (3) to the $\delta = 0$ values. The CO trajectories, agreement counts and expansion in this section are descriptive (Section 4.3). $T_{CO}$ is reported in Section 6.5.

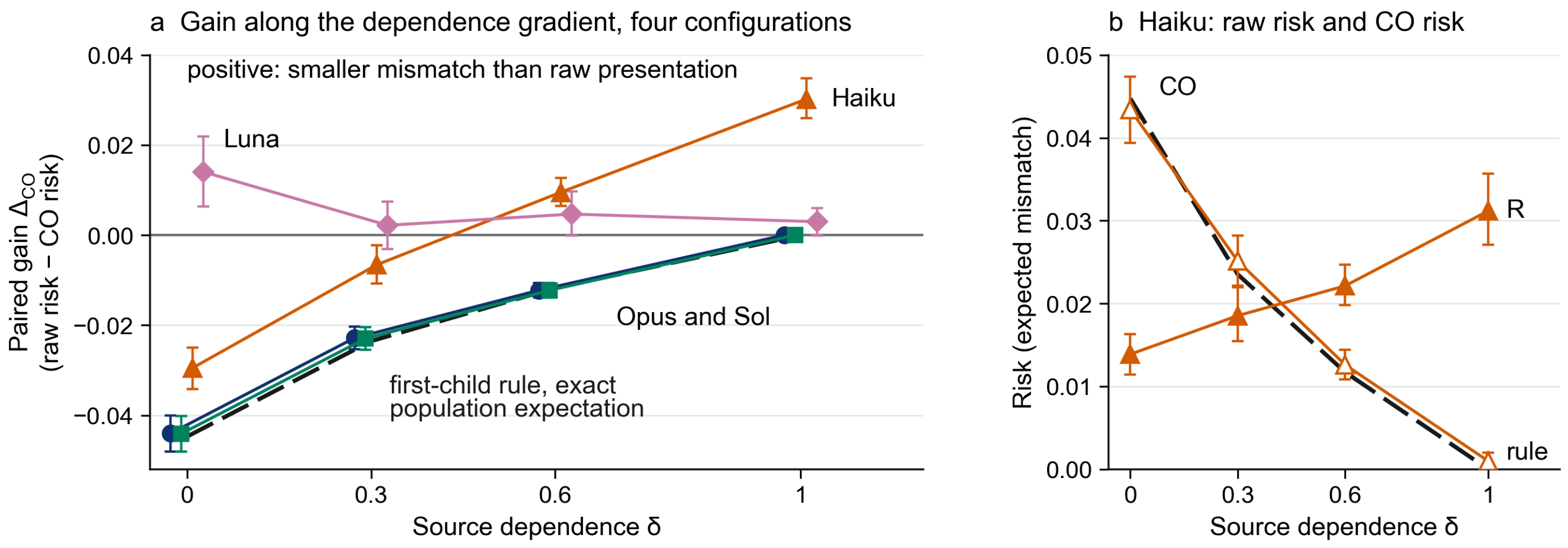

**Fig. 4. Paired gain against the rule's target risk at every dependence level.** (a) Paired gain $\Delta_{CO} = E[(q_R - b)^2 - (q_{CO} - b)^2]$ of the four closed-weight configurations at δ = 0, 0.3, 0.6 and 1. Positive values mean smaller mismatch with the exact posterior than under raw presentation. Points are natural-distribution weighted means with 95% stratified percentile intervals from 10,000 world-level resamples, and lines join the four measured levels. The dashed line is the exact population expectation of the first-child rule's mismatch (Fig. 2c), computed without a model call, plotted as its negative: −0.0447, −0.0235, −0.0117 and 0. It is the population gain of a receiver whose raw answers equal b and whose CO answers equal the rule's output. The realised-world values of Section 3.4 (0.0440754, 0.0229553, 0.0122546 and 0) enter Eq. (3). Opus and Sol follow the line, which lies within their intervals at δ = 0, 0.3 and 0.6. Haiku crosses zero between δ = 0.3 and 0.6. Luna's point estimates are positive at every level, and its interval excludes zero at δ = 0. (b) Haiku: raw risk and CO risk, the expected mismatch $(q - b)^2$ with the exact posterior under raw presentation (R) and the count-once instruction (CO), at each δ, with 95% intervals from the same resampling procedure. R: 0.0139, 0.0185, 0.0221 and 0.0312; CO: 0.0434, 0.0251, 0.0126 and 0.0010. The dashed line is the rule's exact population mismatch of Fig. 2c, 0.0447, 0.0235, 0.0117 and 0, plotted as is, and the CO series follows it. All quantities are descriptive. N = 2,976 worlds.

### 6.1 High agreement with a loss: Opus and Sol

Opus and Sol agreed with the rule's output within 0.001 in all 744 worlds at every δ (2,976 of 2,976). Their raw answers also lay within 0.001 of b: at δ = 0 in 743 and 744 of 744 worlds (2,974 and 2,975 of 2,976 across the panel). For a receiver whose raw answers lie within 0.001 of b, faithfully implementing a rule with positive target risk transfers that risk into the instructed answers. Their paired gains are negative wherever the rule's target risk is positive, −0.0440 [−0.0481, −0.0401] and −0.0441 [−0.0481, −0.0401] at δ = 0, and zero to within $10^{-9}$ at δ = 1, where the rule is exact (Table S4a). The gains track the rule's target risk, which the reference gives without a model call. Both configurations follow the dashed reference line of Fig. 4a, the negative of the rule's exact population mismatch of Fig. 2c (−0.0447, −0.0235, −0.0117 and 0). The line's values at δ = 0, 0.3 and 0.6 lie inside the gain intervals. At δ = 0 the loss is about one third of the Bayes Brier risk of 0.1331.

### 6.2 Low agreement with a gain: Luna

Luna is the opposite pole at δ = 0. Its raw answers lean towards the prior, its most frequent match under R (1,578 of 2,972 valid answers). Luna's raw risk at δ = 0, 0.0789, exceeds the rule's target risk of 0.0441. Under CO, 165 of 744 answers lay within 0.001 of the rule's output and the compliance error was $E[\eta^2] = 0.0341$ (Table S4e). Luna's CO risk fell to 0.0648, and the paired gain was +0.0140 [0.0064, 0.0219] (Fig. 1a, b). Of the 165 agreements, 164 fell in worlds where the rule's output equals the prior. Where the two differ, Luna agreed with the rule's output in 1 of 415 worlds at $\delta = 0$ (0 of 406, 1 of 404 and 2 of 414 at the other levels), against 415 of 415 for Opus and Sol. Luna's gain comes with near-zero agreement wherever the rule's output departs from the prior. Under CO the prior remains Luna's most frequent match, in 1,266 of 2,976 answers against 1,578 of 2,972 under R. The gain is largest at $\delta = 0$, the one level where its interval excludes 0 (Table S4a; Fig. 4a). The dependence-aware note also moves Luna off the prior (1,400 of 2,976 prior matches), and the note's gain at $\delta = 0$ is near zero (Section 6.5).

### 6.3 A reversal with dependence: Haiku

Haiku shows the third pattern, a change of sign with dependence: its gain is −0.0295, −0.0066, +0.0095 and +0.0302 at $\delta = 0$, 0.3, 0.6 and 1, and all four intervals exclude 0 (Table S4a, Fig. 1c). The sign changes between the measured levels $\delta = 0.3$ and 0.6 (Fig. 4a). Haiku's CO answers lay within 0.001 of the rule's output in 727, 647, 702 and 729 of 744 worlds (2,805 of 2,976). Agreement is high and nearly equal at the two endpoints (727 and 729) while the gain changes sign: the sign is carried by the raw baseline and the target risk. Haiku's raw risk rises with $\delta$ (0.0139, 0.0185, 0.0221 and 0.0312) while the realised-world target risk of the rule falls from 0.0441 to 0. Its CO risk is 0.0434, 0.0251, 0.0126 and 0.0010 (Fig. 4b), so its CO answers carry roughly the rule's target risk at every level. The gain is the difference between a raw baseline that worsens with dependence (its raw answers give agreeing sources more than the exact weight at every level; Section 5) and a target risk that falls to zero.

### 6.4 One expansion reads all three patterns

At $\delta = 0$, for Opus and Sol the raw risk is negligible ($3 \times 10^{-5}$ and $3 \times 10^{-14}$) and the squared deviation is below $10^{-9}$, so their gains are approximately the raw risk minus the realised-world target risk (Supplementary Section S2.3). For Haiku the raw risk (0.0139) lies far below the target risk and the squared deviation is small (0.0005), so the gain is negative. For Luna the raw risk (0.0789) exceeds the target risk (0.0441). If Luna implemented the rule faithfully, it would gain the difference, its reference gain. Luna's squared deviation (0.0341) is almost as large as this reference gain. Its CO risk (0.0648) lies below the target risk plus the squared deviation (0.0441 + 0.0341 = 0.0782), so the cross term is negative. On average, Luna's deviations from the rule's output point towards b, which lets Luna keep part of its reference gain. On the 0/1 decision scale the same separation holds for all four receivers (Table S4h, point estimates). At $\delta = 0$ the regret of Opus and Sol rises from $3.9 \times 10^{-18}$ under R to 0.0713 under CO, that of Luna falls from 0.1320 to 0.1104, and that of Haiku rises from 0.0325 to 0.0712. At $\delta = 1$ Haiku's regret falls from 0.0200 to 0.0013.

### 6.5 The four-configuration mean and $T_{CO}$

The four-configuration mean $\Delta_{CO}$ is −0.0259, −0.0126, −0.0026 and +0.0083 at $\delta = 0$, 0.3, 0.6 and 1, and all four 95% intervals exclude 0. The endpoint contrast is $T_{CO} = 0.0342$ [0.0303, 0.0382] (Table 3). The positive $T_{CO}$ comes from three configurations: the per-configuration values are

0.0440 (Opus), 0.0441 (Sol), 0.0597 (Haiku) and −0.0111 [−0.0196, −0.0027] (Luna). Luna's gain is larger at $\delta = 0$ than at $\delta = 1$. On common difficulty support the four-configuration means are −0.0249, −0.0119, −0.0023 and +0.0052, which keep the signs of the unadjusted means (Table S4d). The paired gain protocol puts the two prompts on one scale: the CO-minus-DA mismatch difference is 0.0137 [0.0111, 0.0163] at $\delta = 0.3$ and 0.0043 [0.0026, 0.0061] at $\delta = 0.6$ (Table S4f). The same protocol, run on the dependence-aware note, finds a prompt with near-zero loss: its smallest point gain across the 16 cells of the four closed-weight configurations is −0.0002, against −0.0441 for CO. At four decimals CO gains more than the note in three cells: Luna at $\delta = 0$ (0.0140 against 0.0006) and Haiku at $\delta = 0.6$ (0.0095 against 0.0004) and $\delta = 1$ (0.0302 against 0.0027). A per-receiver evaluation finds them (descriptive point estimates; Table S4c). In the two open-weight configurations (Section 6.6), the prompt that gains more changes with the receiver and δ. CO leads for Qwen at $\delta = 0$, 0.6 and 1 and for Gemma at $\delta = 0.6$ and 1, and the note leads for Qwen at $\delta = 0.3$ and for Gemma at $\delta = 0$ and 0.3 (post hoc descriptive; Table S5d). Prompts are therefore compared receiver by receiver.

### 6.6 Open-weight paired-gain reproduction under a different answer template

Two open-weight configurations, Qwen2.5-32B-Instruct-FP8-dynamic (Qwen) and gemma-3-27b-it-FP8-dynamic (Gemma), answered the revised template P0-draft-3 (Table 2; Section 7.2). In a post hoc descriptive reproduction of the gain pattern (deviation D8; point estimates; Table S5d), Qwen combines low agreement with a positive $\Delta_{CO}$ at every δ, including $\delta = 0$, where its raw risk lies below the rule's target risk. Qwen's CO answers lay within 0.001 of the rule's output in 169 of 2,976 worlds, and its $\Delta_{CO}$ reaches 0.0778 at $\delta = 1$. Gemma's lay within 0.001 in 13 of 2,976 worlds, and its $\Delta_{CO}$ runs from −0.0255 at $\delta = 0$ to +0.0429 at $\delta = 1$. Its gain changes sign with dependence while its agreement is near zero, a further instance of the separation of Sections 6.1–6.3. At $\delta = 1$, where the rule's output equals b, the CO risk in Table S5d is the squared deviation from the rule's output, so the low counts mark departures of measurable size. At $\delta = 0.3$ Gemma pairs a raw risk above the rule's target risk with a negative gain. This cell and Qwen's at $\delta = 0$ are cases in which the receiver's response reverses the sign of the reference gain (Section 8.2).

## 7. Controls: evidence type, reliability level, model family and prompt condition

Three controls test the reach of the separation of Section 6 and the baselines of Section 5. Section 7.1 reports a wording control with new child reports, verbatim copies and paraphrases, Section 7.2 a second reliability level with two open-weight configurations, and Section 7.3 two post-registration prompt conditions, A and B. Per-configuration results are in Tables S5a–S5i and Fig. S1. The main-panel row of Fig. S1c belongs to the confirmatory family. The wording control, the second reliability level and the open-weight configurations are descriptive, and conditions A and B are exploratory.

### 7.1 New child reports against copies and paraphrases

The wording control shows that the partially dependent reports of the main design are new child reports with finite increments. It replaces one report in a fixed slot with a new child report, a verbatim copy or a template paraphrase, at $\delta = 0.3$ and 0.6 over 240 parent worlds (Table S5a, Fig. S1a, b). A new child report has a positive exact increment, and the exact copy-minus-paraphrase difference is 0. Opus and Sol move by 1.00 times the exact shift for a new child report in both arms and at both levels, and they differ by at most 0.0006 between copy and paraphrase. For these two receivers the wording effect is therefore separable from the evidence increment. Haiku's ratios (1.20 to 2.80) are in line with its raw weights above the exact values (Section 5). Luna's ratios (−0.38 to 0.75; all intervals include 0) are in line with its raw lean towards the prior.

### 7.2 A second reliability level and open-weight configurations

The $r = 0.70$ panel carries the raw weighting result to a second reliability level. Run under arms R and G, it keeps the sign of the main panel ($D_R = 0.236$ [0.097, 0.378] against an exact 0.410). The $\nu$ gate passes for Opus, Sol and Haiku and fails for Luna, so this level is read through $D_R$ (Tables S5b and S5c).

The two open-weight configurations returned 14,400 valid answers each (Table 2). They ran under template P0-draft-3 and a 256-visible-token cap, against 4,096 tokens for the closed-weight configurations (deviations D2 and D3). Because prompt formatting shapes the performance of open-weight LLMs [47], the comparison is between configurations as deployed. Like Haiku (Section 5), both give agreeing sources far more weight than the exact value at full dependence ($w3 = 3.377$ and 5.512 against 1, and 2.914 for Haiku). As with Haiku and Luna, their raw source-weight contrasts have intervals that include 0 (Fig. S1c). Their grouping endpoint differences are 0.0345 and −0.0797, and both intervals exclude 0.

### 7.3 Post-registration conditions A and B

Conditions A and B measure how far raw baselines move when the prompt configuration and the run period change together. Section 8.4 draws the unit of evaluation from them. They are exploratory post-result additions (deviations D4, D5 and D7) on 998 main-panel worlds under arms R and G. Each contrast with the main run is a joint difference in prompt and run period, and for Opus and Haiku condition A adds a change of channel (Vertex AI, deviation D7; Supplementary Sections S5.4–S5.5).

Condition A withholds $a$, $e$ and $\delta$ (the generating parameters that the reference discloses by design), gives dependence as one of four ordered verbal levels and keeps $\pi$ and $r$. It probes raw weighting under these withheld parameters. Its reference $b$ is the exact posterior under the full generating information, which the receiver was not shown. Measured against this external reference, $D$ stays positive for Opus and Sol, 0.432 [0.340, 0.521] and 0.516 [0.458, 0.570], against the exact 0.428. The intervals for Haiku and Luna include 0. Under condition A the raw baselines also move: Opus and Sol answer within 0.001 of the full-information posterior in 319 and 435 of 998 worlds, against 998 of 998 under disclosure. Opus weights an agreeing source at $\delta = 0.3$ by 1.857 against the exact 1.702.

Condition B tests the sensitivity of raw baselines to the answer sentence: it gives the closed-weight configurations the answer sentence of the open-weight replication. The paired changes in the four-configuration $T_G$ and $D_R$ from the main run on the same worlds have intervals that include 0. By contrast, Luna's D changes by 0.400 [0.019, 0.799]. Haiku's answers within 0.001 of the exact posterior fall from 375 to 75 of 998, and its mean output length across arms R and G falls from 188.7 to 15.1 tokens.

## 8. Discussion

### 8.1 The information cost of counting once

Under a known evidence model, the price of a count-once prescription is fixed before any model call. The first-child rule is exact at $\delta = 1$ and costs more the weaker the dependence (Section 3.4; Fig. 2c), so idempotence insures against full dependence and charges a premium at every weaker level. For conservative input estimates, covariance intersection guarantees a consistent covariance under unknown cross-correlation [5, 15], and the cautious rule combines non-distinct evidence idempotently [6]. Both give up information to protect against repeated evidence. The first-child rule shares their idempotence, and its own guarantee is exactness at full dependence. Work on partial or unknown dependence [7, 15, 22, 23] extends these operators, and the reference states the price where the dependence is known.

Idempotence settles the treatment of exact repetition. Under partial dependence the price is set by how the rule selects and weights. Root majority and the first-child rule give an agreeing source the same weight, $w_3 = 1$, and differ in the weight of a 2:1 split source, which sets the difference between their prices (Section 3.4). Weighting also separates the Gaussian discount [3, 10] from the binary one (Section 3.1), because dependence maps to weight through the observation model.

Deduplication is exact for zero-increment copies and approximate for new reports from a shared source. Origin-aware deduplication equals b because it joins the removal of copies to the correct root–child likelihood. A new report from a shared source still adds evidence (Section 3.1), and Opus and Sol move by 1.00 times its exact shift (Section 7.1). Deduplication by source label is an engineering shortcut that is exact in the zero-increment regime of the verbatim-repeat evaluations [11, 12, 13, 14].

### 8.2 The receiver is part of the prescription

The value of one prescription is set jointly by the receiver's raw risk and its response to the instruction, so agreement and paired gain are different properties. Agreement counts record how often answers fall at the rule's output. The implementation term of Section 4.1 depends on the magnitude and alignment of deviations under the natural-distribution weights. At $\delta = 0$, where the rule's target risk is largest, agreement ranks the receivers in reverse order of the instruction's value (descriptive). By agreement Opus and Sol (744) lead Haiku (727) and Luna (165 of 744). By paired gain Luna (+0.0140) leads Haiku (−0.0295) and Opus and Sol (−0.044).

Among the four closed-weight configurations, the point estimate of the paired gain has the sign of the reference gain in every cell where the rule or the raw baseline carries risk (post hoc descriptive; Tables S4b, S4c). Opus and Sol, whose raw answers lie within 0.001 of b in nearly every sampled world (Table S3f), have a reference gain close to the negative of the target risk and a negligible implementation term (Section 6.1). Luna keeps part of its reference gain (Section 6.4), and Haiku's reference gain changes sign (Section 6.3). A count-once instruction is designed for receivers that over-count, as Haiku does at every dependence level (Section 5). Across the four closed-weight configurations the sign of the instruction's gain follows the receiver's raw risk relative to the target risk (post hoc descriptive). At $\delta = 0$ Haiku, whose raw risk lies below the target risk, loses 0.0295, and Luna, whose raw answers lean towards the prior and whose raw risk lies above it, gains 0.0140. At $\delta = 1$ Haiku gains 0.0302.

The instructed arm decides whether the gain interval lies above the margin of Section 8.3. For Luna at $\delta = 0.3$, 0.6 and 1 the interval includes 0 (Table S4a), so the adoption rule keeps raw presentation in those strata. In the open-weight reproduction the response to the instruction also gives the gain the opposite sign to the reference gain, for Qwen at $\delta = 0$ and Gemma at $\delta = 0.3$ (Table S5d; post hoc descriptive). Adoption is therefore read from the paired gain of each configuration, and the audit runs both arms.

Zhou and Lu [26] separate the rescue and the harm of decision-level fusion, and the exact posterior puts the benefit and the harm of a fusion instruction on one scale. Krishna et al. [9] and Zhang et al. [28] separate recognising lineage and epistemic cues from using them, and the reference adds that implementing a dependence rule is distinct from fusing well.

### 8.3 From the exact reference to deployment

The paired gain is the improvement in expected Brier score of the instructed answers over the raw answers [48]. By Eq. (1), for the same evidence item x, $E[(q_R - Y)^2 - (q_X - Y)^2 \mid x, q_R, q_X] = (q_R - b(x))^2 - (q_X - b(x))^2$, because Y is conditionally independent of the answers given x (Section 3.2) and the Bayes risk $b(x)(1 - b(x))$ cancels. The Brier difference between the raw and instructed arms on labelled deployment items therefore estimates the paired gain of Eq. (2) on those items without b. A deployment audit sets its interval from its own sampling unit, pairing and clustering structure.

The raw risk bounds the gain of any instruction, because the paired gain is the raw risk minus the instructed risk and a risk is non-negative. The audit for a deployed receiver follows one of two paths, each with its own premise and bound.

1. Reference path. Pipelines that estimate source reliability and dependence upstream and pass the estimates to the fusion node, as truth discovery [18, 19] and dependence-aware data integration [17] do, can screen a receiver before deployment against the posterior those estimates imply.

The screen draws worlds from that evidence model. The path needs no labels, carries no outcome noise and reads the gain in each dependence stratum. Compute the rule's target risk at each dependence level without a model call. It ranks candidate rules and, once the raw arm of the next step has run, gives the reference gain, the gain of a receiver that implements the rule faithfully. Run the raw arm in the production answer format, because raw baselines differed under the condition-B configuration (Section 7.3). Its raw risk bounds the gain. Run the instructed arm and read the paired gain in each dependence stratum of that evidence model, or weight the strata by the expected dependence composition for one figure. The instructed arm measures the implementation term of Eq. (3), so the gain can fall below the reference gain or rise above it. Luna's gain at $\delta = 0.3$, 0.6 and 1 lies far below its positive reference gain, and its interval includes 0 (Tables S4a and S4b). Qwen's reference gain at $\delta = 0$ is negative and its gain is positive (Table S5d; post hoc descriptive).

2. Labelled path. It needs a receiver that reports a probability or a decision for each labelled deployment item. Run the raw and instructed arms on the items. The Brier difference between the arms estimates the paired gain directly, without $\delta$ or an exact posterior. Labels replace b. The raw arm's Brier score, which bounds the gain, includes the Bayes risk (Eq. (1)), and the Brier difference carries outcome noise. For each item it equals the mismatch difference plus $2(b(x) - Y)(q_R - q_X)$, a term with conditional mean 0 and conditional variance $4b(x)(1 - b(x))(q_R - q_X)^2$. An interval of the same width therefore needs more items than on the reference path. For a pipeline that outputs decisions rather than probabilities, the difference in error rates between the arms estimates the difference in 0/1 regret in the same way.

On either path, measure the instruction's per-request input and output token cost from the same two arms. Set an adoption margin in Brier units for that cost. Adopt the instruction when the gain interval lies above this margin. For a nearly deterministic receiver, rare deviations misplace the tails of within-cell percentile intervals (Supplementary Section S3.3). The interval compared with the margin is then the one widened by a deviation budget as in Table S3k. When the interval lies below that margin or reaches it, keep raw presentation. For a receiver whose raw answers already lie near b and that implements the rule, a rule with positive target risk passes that risk into its answers (Section 6.1).

The audit of one receiver on the main panel needs the R and CO arms, 5,952 requests. The full matrix of 14,400 items per configuration contains them. It cost USD 0.43 to 80.06 at Batch prices (Table 2), which caps the audit's cost for each of these receivers. Section 6.5 compares DA with CO on the same scale. A deployment version of the note is a new prompt and is scored again on the labelled path.

### 8.4 Design scope and the next prescriptions to price

The graded-dependence exact reference discloses its generating parameters by design. In fusion, the consistency of estimators is tested in simulations with known truth and known noise statistics, for example with the normalised estimation error squared [49], and disclosure plays the same role here.

Disclosure separates evidence weighting from the estimation of source reliability and correlation, so every gap from b is an inference gap under a fixed evidence model. The reference prices a rule where the evidence model is disclosed. Where it is withheld, as in condition A (exploratory; Section 7.3), adoption follows the labelled path (Section 8.3), which needs no disclosure.

From the sampled worlds alone, without a model call, the reference fixes a loss of 0.0441 at $\delta = 0$ for a receiver that answers at the posterior and implements the rule faithfully. The run shows which receivers pay it, which pay less and which gain. The first-child rule is a keep-earliest policy, and root majority is one vote per source. The reference ranks the two before any model call: root majority has less than half the population mismatch of the first-child rule at $\delta = 0$ (0.0188 against 0.0447), so the next instruction to test is priced. For a receiver that implements a rule faithfully the implementation term vanishes, so the ranking carries over to its gain. Under the first-child rule, Opus, Sol and Haiku have small implementation terms (Sections 6.1 and 6.3). At $\delta = 1$ the rule is exact, and the same instruction helps Haiku, whose raw answers weight agreeing sources above the exact value (Sections 5 and 6.3).

Conditions A and B (Section 7.3; exploratory, deviations D4, D5 and D7) show how far raw baselines move when the prompt configuration changes together with the run period. The unit of evaluation is therefore the receiver together with its prompt configuration, and the paired gain protocol is repeated when any part of it changes. An endpoint without a pinned snapshot can change silently, so the audit reruns both arms on archived worlds or labelled items at a fixed interval.

Six prescriptions are next to price, each a new generator or rule for the same reference. The first is a set of conservative operators: the homogeneous-functional family of Taylor and Bishop [7], the cautious rule [6] and the cautious-adaptive rule [23]. Once each is written as a rule on the visible reports, the reference gives its target risk at every $\delta$. The second is the selection of complementary content within a group [31]. The third is unknown $\delta$ together with an imperfect source map. The estimation error of the dependence, the misattribution of reports to sources, an arrival order that a party or a retrieval score controls (keep-earliest hands a source's vote to its first report) and the cost of the prescription can be priced together. The fourth is the LLM as a source of correlation. Larger and more accurate models make highly correlated errors even when architectures and providers differ [50], so a pipeline that fuses several LLMs carries model-level dependence beyond the document-level source labels used here. The fifth is repeated counting across hops of memory, retrieval and message passing [8, 41]. The sixth is the extension of the reference to other group sizes, conflict patterns and dependence structures.

## 9. Conclusion

Against an exact posterior, the value of a count-once instruction is set by the receiver it is given to, and the reference prices the rule before any model call. At conditional independence, the instruction raised posterior mismatch by 0.044 for two configurations that agreed with the first-child output within 0.001 in all 744 worlds. A third configuration agreed in 165 of those 744 worlds and gained 0.0140, and a fourth reversed from −0.0295 to +0.0302 as dependence rose. At $\delta = 0$, where the rule's target risk is largest, ranking these receivers by agreement with the rule reverses their ranking by paired gain (descriptive). The registered raw source-weight contrast of the four-configuration mean was positive ($D_R = 0.259$ [0.182, 0.337]; Holm control rejected $D_R = 0$), and the grouping endpoint difference was not detected ($T_G = 0.0004$ [−0.0023, 0.0031]). Its interval excludes endpoint differences larger than 0.0031 in magnitude and lies within the registered precision bound of ±0.0046. The graded-dependence exact reference fixes every rule's target risk before any model call, and the paired gain, estimable on labelled deployment items as a Brier difference without the exact posterior, decides adoption for each receiver and prompt condition.

## CRediT authorship contribution statement

Tengfei Shao: Conceptualization, Methodology, Software, Formal analysis, Investigation, Data curation, Visualization, Project administration, Funding acquisition, Writing – original draft, Writing – review & editing. Asma Rezaei: Validation, Writing – review & editing. Ali Reza Kamali: Validation, Writing – review & editing.

## Funding

This work was supported by MEXT Supporting Pioneering Research through AI for 1,000 Discovery challenges Program (SPReAD) Japan Grant Number JPMXP1726306059.

## Declaration of competing interest

The authors declare that they have no known competing financial interests or personal relationships that could have appeared to influence the work reported in this paper.

## Data availability

The study was registered on OSF (permanent identifier https://osf.io/dzgj2/) on 28 October 2026, before the formal run began. The registration is under an approved embargo. The review archive, DEP02_local_reviewer_data_code.zip (195 members, 192 of them scientific files; 228,676,743 bytes), is held in a separate private OSF data project. The archive contains the formal result and final records from the 18 panel-by-configuration logs; the generator, worlds, seeds, prompts and all templates; the calibration and CPU analysis programs with their existing outputs; the original frozen registration-time scientific files alongside the corrected current code; and the manifest of the six configurations described in Supplementary Section S6.4. The programs that read the stored responses are cpu/formal_analysis.py, cpu/explore_analysis.py and cpu/open_weight_all_arms.py, and scenario X04 uses cpu/explore_calibration.py. On acceptance, the archive will be deposited publicly with a DOI.

# Supplementary Material

**Double counting in LLM evidence fusion: exact posteriors separate rule compliance from posterior gain**

Tengfei Shao, Asma Rezaei, Ali Reza Kamali

**Guide.** Against an exact posterior, the value of a count-once instruction is set by the receiver it is given to, and the reference prices the rule before any model call. Start with S4. Table S4a separates agreement and paired gain in both directions. Tables S4b–S4c show that, for the four closed-weight configurations, the paired-gain point estimate has the sign of the reference gain in every cell where the rule or the raw baseline carries risk (post hoc descriptive). A deployment audit (Section 8.3 of the article) runs both arms because the instructed arm is measured rather than forecast. For Luna (gpt-6-luna) at $\delta = 0.3$, 0.6 and 1, the paired gain lies far below the reference gain and its interval includes 0 (Tables S4a and S4b). Two open-weight cells have point estimates of the opposite sign to the reference gain (Table S5d; post hoc descriptive). S1 gives the exact reference and rule prices (Table S1b); S2 prompts, scoring and risk expansion (Eq. (S4)); S3 raw baselines and the registered family, $D_R$ with $T_G$; S5 controls and exploratory conditions A and B; S6 registration, D1–D8 and C1–C6.

**Conventions.** $D_R$ and $T_G$ form the registered confirmatory family and are reported together (Table 3; Section S3.2). The registration lists the following prespecified secondary quantities: $T_{CO}$, the compliance error and the compliance difference; $OD_X$ and $T_M$; $D_G$, $D_{CO}$ and $D_{DA}$, with the ratio changes $\Lambda_X - \Lambda_R$; the $\Delta_X(\delta)$ curves and their standardised versions on common difficulty support; the regression diagnostic and the saturation diagnostic $D_{diag}$; the program-rule comparison; and the wording control, the r = 0.70 panel and the open-weight replication. Holm results within the secondary family are flags. Coverage calibration (Section S3.3) concerns the two confirmatory four-configuration quantities, so the intervals of the wording control, the r = 0.70 panel and the open-weight replication are read as descriptive. The open-weight replication also carries deviations D2 and D3. Per-configuration values, trajectories and agreement counts are descriptive, and conditions A and B are exploratory. Each is labelled where it appears. Intervals are 95% percentile intervals from stratified world-level resamples, unadjusted for multiplicity, unless a table states otherwise. The four-configuration mean weights the four closed-weight configurations equally. Short names follow Table 2 of the article: Opus (claude-opus-5-5), Sol (gpt-6.1-sol), Haiku (claude-haiku-5-5), Luna (gpt-6-luna), Qwen (Qwen2.5-32B-Instruct-FP8-dynamic) and Gemma (gemma-3-27b-it-FP8-dynamic).

## S1. The graded-dependence exact reference

This section supplies the prices the article reads before any model call: exact posteriors, finite same-source increments and the target risk of every program rule (Sections 3.1–3.4).

### S1.1 Generator, exact posterior and finite increments

A world has a hidden state $Y \sim \text{Bernoulli}(\pi)$ and two sources $Z_1$ and $Z_2$ that are independent given Y, each with $P(Z_g = Y) = a$. Each source emits three child reports $X_j = Z_g \oplus E_j$, with flips $E_j \sim \text{Bernoulli}(e)$ independent given the sources. The single-report reliability is $r = P(X_j = Y) = a(1 - e) + (1 - a)e$. Two child reports of one source are both correct with probability $a(1 - e)^2 + (1 - a)e^2 = r - e + e^2$. The dependence $\delta$ is the correlation between their correctness, so $r - e + e^2 = r^2 + \delta \cdot r(1 - r)$. The root of this quadratic with $e \leq 1/2$, together with the corresponding a, is

$$e = \frac{1-\sqrt{1-4(1-\delta)\,r(1-r)}}{2}, \qquad a = \frac{r-e}{1-2e}. \tag{S1}$$

At δ = 0, Eq. (S1) gives e = 1 − r and a = 1, so each source equals Y. At δ = 1 it gives e = 0 and a = r.

For a visible pattern x of the six reports with the source map shown in the prompt, let $P(Z_g = z \mid Y = y)$ equal a if z = y and 1 − a otherwise, and let $P(X_j = x_j \mid Z_g = z)$ equal 1 − e if $x_j = z$ and e otherwise. The likelihood of state y with each source summed out, and the exact posterior, are

$$L_y(x) = \prod_{g=1}^{2} \sum_{z\in\{0,1\}} P(Z_g = z \mid Y = y) \prod_{j\in g} P(X_j = x_j \mid Z_g = z),$$
$$b = \frac{\pi\, L_1(x)}{\pi\, L_1(x) + (1-\pi)\, L_0(x)}. \qquad \text{(S2)}$$

All 64 report patterns are enumerated. At δ = 1, patterns in which reports of one source disagree have probability zero, so later reports of a source add nothing. They still keep distinct report IDs and, under wording frame L1, distinct sentences. A copy or a paraphrase of a child report carries no new noise: it enters the likelihood through its origin and adds no evidence. Table S1a lists the parameters and the exact second- and third-report increments at both reliability levels.

**Table S1a.** Generator parameters along the dependence gradient

| δ | a | e | Second-report increment | Third-report increment |
|---|---|---|---|---|
| **Reliability r = 0.65** | | | | |
| 0.0 | 1.0000 | 0.3500 | 1.000 | 1.000 |
| 0.3 | 0.7490 | 0.1988 | 0.527 | 0.176 |
| 0.6 | 0.6881 | 0.1013 | 0.243 | 0.031 |
| 1.0 | 0.6500 | 0.0000 | 0.000 | 0.000 |
| **Reliability r = 0.70** | | | | |
| 0.0 | 1.0000 | 0.3000 | 1.000 | 1.000 |
| 0.3 | 0.8116 | 0.1791 | 0.516 | 0.158 |
| 0.6 | 0.7454 | 0.0926 | 0.237 | 0.028 |
| 1.0 | 0.7000 | 0.0000 | 0.000 | 0.000 |

Notes: Increments are those of the second and third agreeing child reports of one source, in units of one independent report.

Section 3.2 lists what the prompt discloses, and Section S2.1 shows a complete rendered prompt.

## S1.2 Program rules and their exact target risks

Table S1b prices the six program rules at r = 0.65 before any model call. With r fixed, the Bayes risk is 0.1331, 0.1543, 0.1661 and 0.1778 at δ = 0, 0.3, 0.6 and 1 (Section 3.1).

**Table S1b.** Exact reference and program rules at r = 0.65

| | w3 | | | | | v | | | Risk | | | |
|---|---|---|---|---|---|---|---|---|---|---|---|---|
| **Rule** | **δ = 0** | **δ = 0.3** | **δ = 0.6** | **δ = 1** | **w2 (δ < 1)** | **δ = 0.3** | **δ = 0.6** | **D** | **δ = 0** | **δ = 0.3** | **δ = 0.6** | **δ = 1** |
| Exact posterior (= origin-aware deduplication) | 3.000 | 1.702 | 1.274 | 1.000 | 1.000 | 0.630 | 0.560 | 0.428 | 0.0000 | 0.0000 | 0.0000 | 0.0000 |
| Full count | 3.000 | 3.000 | 3.000 | 3.000 | 1.000 | 0.750 | 0.750 | 0.000 | 0.0000 | 0.0092 | 0.0209 | 0.0296 |
| First child | 1.000 | 1.000 | 1.000 | 1.000 | 0.333 | 0.750 | 0.750 | 0.000 | 0.0447 | 0.0235 | 0.0117 | 0.0000 |
| Root majority | 1.000 | 1.000 | 1.000 | 1.000 | 1.000 | 0.500 | 0.500 | 0.000 | 0.0188 | 0.0052 | 0.0012 | 0.0000 |
| Literal-only deduplication | 1.670 | 1.670 | 1.670 | 1.750 | 0.487 | 0.774 | 0.774 | 0.000 | 0.0343 | 0.0258 | 0.0233 | 0.0182 |
| Overlap-factor discount | 0.600 | 0.600 | 0.600 | 0.600 | 0.200 | 0.750 | 0.750 | 0.000 | 0.0418 | 0.0228 | 0.0131 | 0.0043 |

Notes: Source weights w3 and w2 are in units of one independent report (logit r). The weight w2 is undefined at δ = 1, where no source splits. Every cell at δ = 1 is unanimous (Section S2.4), so the w3 of literal-only deduplication moves from 1.670 to 1.750. Risk is the natural-distribution mean of the mismatch $(q - b)^2$, with layers weighted equally. For a rule it is the target risk.

The six rules are computed on the visible information of each world and form a fixed scale on which the arms are placed (Fig. 2c draws their mismatch at $r = 0.65$).

1. **Full count.** Every distinct child report is treated as independent, with reliability r.
2. **First child.** For each source, the earliest child report is counted once with reliability r. This is the procedural version of arm CO (Section S2.1).
3. **Root majority.** For each source, the majority of its three reports is counted once with reliability r.
4. **Literal-only deduplication.** Statements stripped of report IDs, source labels, senders and positions are compared as strings across sources, and the first occurrence of each is kept and treated as independent with reliability r. Under L0 at most two reports remain, one per value, and the mismatch at $\delta = 0$ is 0.0685. Under L1 every line differs, the rule equals full count and the mismatch at $\delta = 0$ is 0.0000. Origin IDs are ignored. In the wording control, the rule acts on the version that contains the copy or the paraphrase.
5. **Origin-aware deduplication.** Copies and paraphrases are removed by origin ID, and the correct root–child likelihood is applied. Its output equals the exact posterior b.
6. **Overlap-factor discount.** The rule applies the overlap factor $1/(1 + \lambda o)$ of RIDE-CAL with $\lambda = 2$ [29] within this study's source partition and aggregation. Each distinct child report is a unit whose source set is its source, and its overlap o sums the Jaccard overlaps with the other units (1 within a source, 0 across sources). Every report therefore has $o = 2$ and adds 1/5 of a report with reliability r on the logit scale.

**Population value and realised-world value of the first-child rule.** Table S1b and Fig. 2c give the population value, the exact expectation over the natural cell distribution and the arrival order, and hence over each source's earliest report. The realised-world value is the target risk of the same rule on the 2,976 realised main-panel worlds, with natural-distribution cell weights and the earliest reports that their sampled arrival orders produced: 0.0440754, 0.0229553, 0.0122546 and 0 at $\delta = 0$, 0.3, 0.6 and 1. Eq. (S4) and Eq. (3) use it (Section 3.4).

## S1.3 Wording frames, cell quotas and slot construction

**Wording frames.** Each child report is a sentence such as "The state is 1." All arms of a world share its wording frame, L0 or L1 (Section 4.1).

**Cell quotas.** Given $\delta$ and $\pi$, a cell (Section 4.1) determines the exact posterior and all evidential content of the prompt. Below $\delta = 1$ each prior has 10 cells. At $\delta = 1$ the cells are the 3 in which each source is unanimous. Within a cell, the assignment of counts to sources, the order of reports within a source, the arrival order, the IDs, the senders and the source labels are exchangeable. In each $\delta \times \pi \times L$ layer, every cell with positive probability receives at least 3 worlds, and the remaining worlds are allocated by the largest-remainder method on the exact cell probabilities. The presentation within a cell is drawn from the generator and Y from the cell's exact posterior, so the joint law given the cell equals that of the generator. In the main panel, the smallest positive cell holds 4 worlds ($\delta = 0.6$, the cell (2,2) at $\pi = 0.25$ and the cell (1,1) at $\pi = 0.75$).

**Slot construction.** In each selected world of the wording control (Section 7.1; Section S5.1), one source is chosen at random and its third child report is replaced. The main version fills the slot with a new child report that carries its own noise, the copy-slot version with a verbatim copy of that source's second report, and the paraphrase-slot version with a template paraphrase of it. The copy and the paraphrase carry the origin ID of the second report and state its value. Every version shows six reports, and in the copy-slot and paraphrase-slot versions the source keeps two distinct child reports. The exact posterior of each version is computed separately. New minus copy measures the weight a receiver gives to a new child report, whose exact increment is positive. Copy minus paraphrase measures wording alone, whose exact difference is zero. Wording-control worlds are answered under arms R and G.

### S1.4 Symbols

Table S1c lists the main symbols, each defined where it first appears.

**Table S1c.** Main symbols

| Symbol | Definition |
|---|---|
| Y, π | Hidden binary state of a world; prior P(Y = 1), stated in the prompt |
| r, δ | Single-report reliability $P(X_j = Y)$; dependence, the correlation between the correctness of two child reports of one source |
| a, e | Source reliability $P(Z_g = Y)$; child noise, the flip probability of a child report given its source |
| b | Exact posterior from the information the prompt shows |
| R, G, CO, DA | Arms: raw presentation, grouping by source, the count-once instruction, dependence-aware note |
| q, ε | Parsed reported probability, including the flagged prior fallback; truncation scale of logit q |
| d, $\Delta_X(\delta)$ | Paired difference $d = (q_R - b)^2 - (q_X - b)^2$; its natural-distribution mean $\Delta_X(\delta)$, the paired gain of arm X ∈ {G, CO, DA} |
| $q_{FC}$, η | Output of the procedural first-child rule; deviation of the CO answer from it, $q_{CO} - q_{FC}$ |
| w3, w2, α | Projected weights of an agreeing and of a split source, in units of one independent report; prior coefficient |
| D, $D_R$, $D_X$ | Source-weight contrast; its four-configuration mean in arm R; D in the other arms |
| $T_G$ | Endpoint difference of the grouping effect |
| v, Λ | w3/(w3 + w2); v(0.3) − v(0.6) |
| $T_{CO}$ | Endpoint contrast of the first-child rule, $\Delta_{CO}(1) - \Delta_{CO}(0)$ |

Notes: In condition A, b is the external full-information reference (Section 7.3). $\Delta_G$ is the symbol used throughout. The registration used $\Delta_C$ for it. In the rendered prompt text (Section S2.1), the dependence parameter δ is written as d. Everywhere else d is the paired difference. Symbols local to one section ($R_{raw}$, $R_{CO}$, $R_{FC}$, $D_{diag}$, β, γ, K) are defined there.

## S2. The paired gain protocol

This section fixes the measurement and the registered rules: the prompts of the four arms (raw presentation R, grouping by source G, the count-once instruction CO and the dependence-aware note DA), the mismatch score and the risk expansion of Eq. (3) (Sections 3.2, 3.3 and 4), then the projection, estimands, bootstrap and result branches under which Section 5 reports $D_R$ with $T_G$.

### S2.1 Prompts and presentation arms

The R and G prompts of a world have identical character counts across all 4,512 worlds of the formal, pilot and admission panels, and identical token counts under both the Qwen2.5 and the gemma-3 tokenizers. G is a reversible regrouping that keeps conflicts and multiplicity and adds no posterior, formula or answer. Under the Qwen2.5 tokenizer, a main-panel request averages 552 tokens for R and G, 586 for CO and 598 for DA, chat template included. The CO and DA sentences are the same in every world and wording frame.

The prompt below was rendered with the production renderer and the study templates for one world at r = 0.65, δ = 0.3, π = 0.5 and wording frame L0: source S-F has three reports saying 1, source S-B is split 2:1, and the prompt text writes δ as d. The other arms and conditions differ from it in the lines given below.

#### Arm R

```
Task: estimate the probability that a hidden binary state equals 1.

How the evidence was generated:
1. The state is 1 with prior probability 0.50 and 0 otherwise.
2. There are two sources. Each source has its own reading of the state. Given the state, the two
readings are independent, and each equals the state with probability a = 0.7490 (rounded to four
decimals).
3. Each line below belongs to one source. An original report states its source's reading, flipped
with probability e = 0.1988 (rounded to four decimals); flips are independent across original
reports given the readings.
4. A line whose origin ID differs from its own report ID is a verbatim copy or a rewording of the
```

```
earlier report with that ID. It states the same value as that report and adds no new noise.
5. Taken alone, each original report is correct with probability r = 0.65. For two original
reports from the same source, the correlation between their being correct is d = 0.30. The exact
a and e are the values implied by r and d.
6. Each source contributes three lines. Arrival numbers, report IDs and senders carry no
information about the state.

Reports are listed by arrival time.
Source S-F: three lines
Source S-B: three lines
#1 | report R-628 | origin R-628 | sender Agent 33 | source S-F | The state is 1.
#2 | report R-673 | origin R-673 | sender Agent 56 | source S-B | The state is 1.
#3 | report R-869 | origin R-869 | sender Agent 18 | source S-F | The state is 1.
#4 | report R-439 | origin R-439 | sender Agent 55 | source S-B | The state is 0.
#5 | report R-543 | origin R-543 | sender Agent 14 | source S-F | The state is 1.
#6 | report R-914 | origin R-914 | sender Agent 47 | source S-B | The state is 1.

Answer with only a JSON object of the form {"p_state_1": p}, where p is your probability (between
0 and 1) that the state is 1.
```

Arm G replaces the line "Reports are listed by arrival time." with "Reports are listed by their source." and lists #1, #3 and #5 under Source S-F and #2, #4 and #6 under Source S-B, each report line unchanged.

Arms CO and DA each add one paragraph before the answer instruction (CO the first, DA the second). Under CO all six reports remain in the prompt.

```
Rule: for each source, use only its earliest original report and count it once as a report that
is correct with probability r; ignore that source's other lines.

Note: a copy or rewording of a report adds no new information. The contribution of different
original reports from the same source is determined by the generating process above; avoid both
over-counting and under-counting them.
```

## S2.2 Parsing, outcomes and fallback

The parser takes the first fragment of the visible text that decodes as a whole JSON object with the key p_state_1. Its value is the scored q and must be a number in [0, 1]. One main-panel answer held more than one distinct value (Haiku, arm CO, with the first and last values 0.2347 apart). Semantic failures (unparseable output, values outside [0, 1], non-numeric values, refusals, truncations and empty output) are scored at the prior $\pi$ with a flag and stay in every estimate without replay.

Transport failures (network errors, timeouts, server errors and, in Batch results, errored, cancelled or expired items) receive at most one replay with the same input, while their block is running and the panel's replay reserve has units left. The reserve is 5% per configuration and panel: 595 for the main panel, 77 for the r = 0.70 panel and 48 for the wording-control panel. A replay that fails again is treated as a semantic failure and flagged as a transport failure. Rejected requests (invalid-request errors, item-level client errors and batches that fail validation) stop the block, and rate-limit responses are resent after the indicated wait outside the failure and replay counts.

On the main panel, the largest gap in failure rate between two arms within a $\delta$ level is 0.0027 for Luna (overall failure rate 0.0006), within the one-percentage-point limit for comparing arms. Across the three panels, Luna has 11 final failures with the flagged prior fill: nine refusals under deviation D1 (Table S2a; Section S6.2.1) and two truncations (main panel, arm G at $\delta = 0$; r = 0.70 panel, arm G). The complete-case version of the confirmatory family is a descriptive check on the 2,969 worlds in which all eight R and G answers of the four configurations are valid, aggregated with the original cell probabilities (Table S3h).

**Table S2a.** Items recorded as refusals under deviation D1 (world, version, arm)

| Panel | Configuration | World | Version | Arm |
|---|---|---|---|---|
| formal_main | gpt-6-luna | formal_main-r65-d6-p50-L1-0036 | main | G |
| formal_main | gpt-6-luna | formal_main-r65-d6-p75-L1-0063 | main | R |
| formal_main | gpt-6-luna | formal_main-r65-d6-p75-L1-0094 | main | R |
| formal_main | gpt-6-luna | formal_main-r65-d6-p25-L0-0058 | main | G |
| formal_main | gpt-6-luna | formal_main-r65-d3-p50-L0-0115 | main | R |
| formal_main | gpt-6-luna | formal_main-r65-d3-p75-L1-0012 | main | R |
| formal_wording | gpt-6-luna | formal_main-r65-d3-p50-L1-0001-paraphrase | paraphrase_slot | R |
| formal_wording | gpt-6-luna | formal_main-r65-d6-p50-L0-0006-copy | copy_slot | G |
| formal_wording | gpt-6-luna | formal_main-r65-d6-p50-L1-0000-copy | copy_slot | R |

## S2.3 Scoring, weighting and the risk expansion

The score is the mismatch $(q - b)^2$ of main Eq. (1), with q the parsed reported probability, including the flagged prior fallback. For an arm X compared with R, the paired difference of one world and its natural-distribution mean at dependence δ are

$$d = (q_R - b)^2 - (q_X - b)^2, \qquad \Delta_X(\delta) = E_\delta[(q_R - b)^2 - (q_X - b)^2], \tag{S3}$$

Positive values mean smaller mismatch than R (main Eq. (2)). The 0/1 decision regret is scored for the decision that chooses 1 when $q \geq 0.5$: it equals $|2b - 1|$ when that decision differs from the Bayes decision with the same information, and 0 otherwise. Mismatch and regret use the raw q, and logit quantities use logit_clip(q), which truncates q to $[\varepsilon, 1 - \varepsilon]$.

**Natural-distribution weighting.** Cell and layer weights are those of Section 4.1, and four-configuration means are taken after the layer means.

**Truncation.** Every exact posterior lies inside $[\varepsilon, 1 - \varepsilon]$ at $\varepsilon = 0.001$ (the most extreme is about 0.00206 or 0.99794). The confirmatory family, the result branches and all closeness wording are computed at $\varepsilon = 0.001$, and $\varepsilon = 10^{-6}$ is a prespecified sensitivity analysis on the same resamples (Tables S3h and S3i). The five registered scale checks are the sign of $D_R$, its Holm decision, every configuration's ν gate in every arm, the direction supported by the four-configuration mean Λ interval and the result branch (Section S3.2).

**Difficulty standardisation.** Difficulty bands are fixed by b before treatment: the middle band $b \in [0.4, 0.6]$, the inner band (the rest of [0.1, 0.9]) and the outer band (outside [0.1, 0.9]). Standardised comparisons across δ (Table S4d) use, for each prior, the bands with at least 5% natural mass at all four δ, weighted by their mean natural mass. At r = 0.65 these are the inner and outer bands for the priors 0.25 and 0.75 (weights 0.533 and 0.467) and the middle and inner bands for the prior 0.5 (0.365 and 0.635). The natural-distribution estimates are primary.

**Risk expansion for arm CO.** Let $R_{raw}$, $R_{CO}$ and $R_{FC}$ be the natural-distribution mean mismatches $E[(q_R - b)^2]$, $E[(q_{CO} - b)^2]$ and $E[(q_{FC} - b)^2]$, where $q_{FC}$ is the output of the procedural first-child rule on the same realised worlds. Let $\eta = q_{CO} - q_{FC}$ be the deviation of the answer from the rule's output. Writing $q_{CO} - b = (q_{FC} - b) + \eta$ and expanding the square gives, world by world and hence in the weighted mean,

$$\begin{gathered} (q_{CO} - b)^2 = (q_{FC} - b)^2 + \eta^2 + 2\,\eta\,(q_{FC} - b), \\ R_{CO} = R_{FC} + E[\eta^2] + 2\,E[\eta\,(q_{FC} - b)], \\ \Delta_{CO} = R_{raw} - R_{CO}. \end{gathered} \tag{S4}$$

Eq. (S4) restates main Eq. (3) per world and defines $\Delta_{CO}$. The squared deviation $E[\eta^2]$ is the compliance error, and $R_{FC}$ is the realised-world target risk of Section S1.2. For Opus and Sol, $E[\eta^2]$ is below $10^{-9}$ at every δ (Table S4e), so $\Delta_{CO} \approx R_{raw} - R_{FC}$.

**Agreement.** Agreement with a rule's output means $|q - q_{rule}| \le 0.001$, with the absolute difference rounded to 12 decimals before the comparison (code correction C6, Section S6.3), so a difference of exactly 0.001 counts as agreement. Agreement describes the answer, whatever process produced it. Table S4a (Section S4.1) gives the counts by dependence level.

## S2.4 Projection, estimands and diagnostics

**Fixed common projection.** For each configuration, arm and δ, the cell-mean responses are summarised by the projection of main Eq. (4),

$$\operatorname{logit} q \approx \alpha \operatorname{logit} \pi + \operatorname{logit} r \, (w_3 U + w_2 S), \tag{S5}$$

with U, S, w3 and w2 as in main Section 4.3. The weights are fixed in advance and shared by both middle levels, so a difference between levels reflects responses on one fixed set of cells. The exact posterior satisfies the form with α = 1, w2 = 1 and the w3 of Table S1b.

Over the 30 cells formed by 3 priors and 10 count pairs, let $m_c(\delta)$ be the mean logit_clip(q) in cell c (the two wording frames weighted equally), $x_c = (\operatorname{logit} \pi, U_c, S_c)$, and $w_c$ the mean of the exact cell probabilities at δ = 0.3 and 0.6, divided by 3. Then $\beta_\delta = G^{-1} \Sigma_c w_c x_c m_c(\delta)$ with $G = \Sigma_c w_c x_c x_c^T$, and w3 and w2 are the U and S coefficients divided by logit r. The same G and weights serve both middle levels and δ = 0. At δ = 1 the cells are the three unanimous ones, so no source splits and w2 is undefined. The w3 at δ = 1 comes from a separate regression on (logit π, U) weighted by the δ = 1 cell probabilities and is compared with the exact value at that level. Table S3e gives the weighted RMS residuals of every projection.

**Confirmatory contrasts and their null receivers.** With $\Delta_G(\delta)$ from Eq. (S3), the two confirmatory quantities are

$$\begin{gathered} D = w_3(0.3)\, w_2(0.6) - w_3(0.6)\, w_2(0.3), \\ T_G = \Delta_G(1) - \Delta_G(0), \end{gathered} \tag{S6}$$

which restate main Eq. (5) and Eq. (6). D is computed per configuration from the projected weights. $D_R$ is an equal-weight mean of D over the four configurations in arm R, and $T_G$ is averaged over the four configurations.

The hypothesis for $D_R$ is H0: $D_R = 0$. Population D is exactly 0 for any receiver that ignores δ, whatever the shape of its response, and for one whose response to δ is a common multiplicative rescaling of δ-independent source weights. Rescaling weights that themselves vary with δ gives $D = c(0.3)c(0.6)D_B$, so Bayes × 0.5 gives 0.107. A receiver that changes its overall conservatism with δ and then hits a fixed cap can have $D \neq 0$: the receiver defined by $\operatorname{logit} q = \operatorname{clip}(\operatorname{logit} \pi + s_\delta \cdot c \cdot N, \pm\log 9)$, with N = 3U + S, $s_{0.3} = 0.5$ and $s_{0.6} = 1$, has population D = 0.077. The saturation diagnostic (last item below) separates this case from dependence-specific weighting. The hypothesis for $T_G$ is H0: $T_G = 0$, and $T_G$ is read as the endpoint difference of the grouping effect. In the planning scenarios $T_G$ is 0 when R = G = Bayes and 0.074 for two fixed rules that ignore δ.

**The ν gate and Fieller-type sets.** The ratio ν = w3/(w3 + w2) and its change Λ = ν(0.3) − ν(0.6) are descriptive. A configuration–arm pair passes the ν gate when three conditions hold at both middle levels: the 95% intervals of w2 and w3 + w2 avoid $[-10^{-9}, 10^{-9}]$, the point estimates of w3 + w2 have the same sign, and these point estimates lie outside the zero band. The gate sets finite-sample reporting conditions for the descriptive ratios. The exact posterior gives ν = 0.630 at δ = 0.3 and 0.560 at δ = 0.6, so $\Lambda_B = 0.070$. Per configuration, ν and Λ have percentile intervals where the gate passes and Fieller-type sets in all cases (Table S3g). The set for Λ inverts the bootstrap sign test of $g(\lambda_o) = a(0.3)s(0.6) - a(0.6)s(0.3) - \lambda_o s(0.3)s(0.6)$, with a = w3 and s = w3 + w2. $D_R$ uses all four configurations and carries no gate.

**Prespecified secondary quantities and diagnostics.** Section S4.2 reports the secondary quantities and defines $OD_X(\delta)$ and $T_M(\delta)$.

1. **Arm CO.** $T_{CO} = \Delta_{CO}(1) - \Delta_{CO}(0)$. The compliance difference $E[(q_R - q_{FC})^2] - E[(q_{CO} - q_{FC})^2]$ is positive when CO is closer to the rule's output than R is.
2. **D for the other arms** ($D_G$, $D_{CO}$, $D_{DA}$). D scales with the square of the evidence scale: R = Bayes and G = Bayes × 0.5 have identical ν, yet $D_G - D_R = -0.321$. Whether grouping or the note changes the ratio is therefore read from $\Lambda_X - \Lambda_R$, where the ν gate passes for all four configurations in both arms.
3. **Regression diagnostic**, per configuration, with one row per world and arm: logit $q = a_0 + a_p$ logit $\pi + \beta$ $LLR_B + \gamma(LLR_N - LLR_B)$, plus arm terms (arm shifts and arm interactions with both slopes, R being the reference arm) and an L1 shift, where $LLR_B$ = logit b − logit π, $LLR_N$ = logit r · N and N = 3U + S. Weights restore the natural cell distribution, and standard errors are clustered by world (CR1). The references are β = 1 with γ = 0 for the exact posterior and β = 1 with γ = 1 for full count. A 95% interval of γ that touches exactly one of the zones [−0.05, 0.05] (exact-reference zone) and [0.95, 1.05] (full-count zone) reads as that type, one that touches neither as other, and one that touches both as undecided. A type means compatibility with one reference and exclusion of the other.
4. **Wording control.** On clipped logits, each world, configuration and arm R or G has the oriented difference $z = (2x - 1)\cdot[\text{logit } q(A) - \text{logit } q(B)]/\text{logit } r$. New minus copy takes $x = x_{new}$, the value of the replaced child report, with A the main version and B the copy-slot version, and is compared with its exact increment. Copy minus paraphrase takes $x = x_{slot}$, the slot value, with A the copy-slot version and B the paraphrase-slot version, and has exact value 0. The slot value orients the differences because undirected differences average to zero over the three symmetric priors. Within each layer the mean is a Horvitz–Thompson estimate over the first 20 worlds of the layer, an equal-probability subsample because the order within a main-panel layer is a random permutation (120 worlds per δ and 480 wording-control versions). Layers are weighted equally.
5. **Saturation diagnostic.** D is recomputed on a fixed set of low-evidence cells, π = 0.5 with (0,2), (0,3), (1,1), (1,2), (1,3) and (2,2), where |N| ≤ 2, using a regression on U and S with the common weights restricted to the set. On this set N takes the values −2, 0 and 2, and the cell probabilities are symmetric. Every receiver that acts on the reports through N alone, with any δ-dependent shape and any cap (including the scale-then-cap example above), therefore projects to a 3:1 ratio of U to S weights and has $D_{diag} = 0$, whereas the exact posterior keeps $D_{diag} = 0.428$. A wide interval of $D_{diag}$ is reported as inconclusive.

## S2.5 Bootstrap, planning basis and registered wording

**Stratified world bootstrap.** The unit of inference is the world. Intervals and tests come from a world bootstrap stratified by δ × π × L × cell, with B = 10,000 resamples, 95% percentile intervals and Holm's step-down procedure [46] on the two p values. Two 97.5% percentile intervals target nominal simultaneous 95% coverage for the pair by Bonferroni adjustment. Within each stratum, $n_s$ worlds are drawn with replacement, and the configurations and arms of a world move with it. The quota rule guarantees at least 3 worlds per stratum (at least 4 in the main panel). The ordinary bootstrap understates the variance of a stratum mean by the factor $(n_s - 1)/n_s$, so the deviation of each resampled stratum mean from the observed mean is multiplied by $\sqrt{n_s/(n_s - 1)}$. The rescaled quantity is each configuration's stratum mean of logit_clip($q_R$) for D, the stratum mean of the panel paired difference d for $T_G$, and the stratum mean of the score for every other quantity. Because the rescaling enlarges the resampled deviations around an observed mean near 0, percentile lower limits of non-negative scores such as squared errors can fall slightly below 0 (Tables S4b and S4e). Each resample recomputes the cell means and the fixed projection, and the two-sided p value is $\min\{1, 2\cdot\min[\#(\theta^* \leq 0), \#(\theta^* \geq 0)]/B\}$. Strata with zero variance, for example when every answer of a stratum is a fallback, enter as usual.

**Planning basis.** The main panel holds 744 worlds per δ, fixed on 22 October 2026 by planning simulations with specified receivers. For $T_G$, the target was a 95% half-width of 25% of the endpoint difference in the scenario where G tracks dependence (approximately 0.0184), which sets the precision bound at approximately ±0.0046. With independent worlds, a standard deviation 1.5 times the planning value and conservative 97.5% intervals, 719 worlds per δ are needed. The cell-quota draw inflates the variance of natural-distribution quantities by at most 1.033, which gives 743, or 744 after rounding to 124 worlds per π × L layer. For $D_R$, recovery simulations used 720 worlds per δ, four configurations sharing worlds, 30% of the logit noise common across configurations and the rescaled stratified bootstrap.

**Result branches and wording rules.** The precision bound for $T_G$ is ±0.0046 (above), and the equivalence bound for Λ is ±0.0174, 25% of $\Lambda_B$. Both are registered reporting bounds. After the Holm decisions, three steps place each result in one branch. The observed branch is reported in Section S3.2.

*Step 1: ratio change under raw presentation ($D_R$).* A ratio direction needs the ν gate and $\Lambda_R$ in addition to the sign of $D_R$. The ratio direction is supported when the ν gate passes for all four configurations in arm R and the 95% interval of the four-configuration mean $\Lambda_R$ excludes 0 with the sign of $D_R$. With the ratio direction supported, the text states that the ratio changes with δ. Otherwise the text states the sign of $D_R$ alone. If the 95% interval of $D_{diag}$ contains 0 or the diagnostic is unavailable, the text adds that overall conservatism followed by saturation remains a possible explanation.

- **A+**: $D_R$ rejected and positive. With the ratio direction supported, the text states that under raw presentation the projected weight ratio of agreeing to split sources changes with δ in the direction of dependence adaptation. If the four-configuration mean $\Lambda_R$ interval excludes 0 with the opposite sign, the text states that $D_R$ is positive and reports the conflict.
- **A−**: $D_R$ rejected and negative. With the ratio direction supported, the text states that the ratio changes in the reverse direction: stronger dependence gives agreeing sources more relative weight.
- **E**: $D_R$ not rejected, all four ν gates pass and the four-configuration mean $\Lambda_R$ interval lies within ±0.0174. The reading is that the ratio under raw presentation stays within the equivalence bound across the two middle levels.
- **U**: all remaining cases, reported as inconclusive. A failed ν gate affects this step alone and leaves an established A+ or A− in place.

*Step 2: grouping and the note.* Statements about grouping and the note require that the interval of $\Lambda_X - \Lambda_R$ excludes 0. $T_G$ is reported with direction and interval as the endpoint difference of the grouping effect. "Grouping changed the ratio" also requires the ν gate to pass for all four configurations in R and G, and the statement is marked as secondary. A detected $T_G$ with an undetected $\Lambda_G - \Lambda_R$ is not attributed to a fixed rule or to general conservatism, because the change may lie at other δ or in the projection residuals.

*Step 3: closeness wording.* "Projected evidence weights close to Bayes" requires the 95% intervals of the four-configuration mean $w2_X$ and of $w3_X(\delta)/w3_B(\delta)$ at δ = 0.3 and 0.6 to lie in [0.75, 1.25]. "Near Bayes" or "calibrated" has two further requirements: the upper 95% limit of natural-distribution mismatch is at most 25% of the full-count mismatch at the same δ (0.0023 at δ = 0.3 and 0.0052 at δ = 0.6), and the $T_G$ interval lies within ±0.0046. This conclusion refers to the two middle levels and the confirmatory panel and means at least a 75% reduction of mean squared deviation relative to full count. Two receivers fail this condition: one equal to Bayes × 0.5 in both arms, and one that adds a constant 1 to logit b in both arms. These phrases are registered wording conditions for the four-configuration panel.

## S3. Raw baselines and the registered family

This section gives each receiver's raw baseline, which, with the rule's target risk, sets the reference gain under faithful implementation. It also reports the registered family $D_R$ and $T_G$ in full, with its calibration record and deviation envelopes (Section 5).

### S3.1 Raw source weights and output references

Tables S3a–S3e and S3g show arm R. Complete values for all arms are in results/formal_analysis.json in the review archive.

**Table S3a.** Source weight w3 of an agreeing source (3:0), arm R.

| Configuration | δ = 0 | δ = 0.3 | δ = 0.6 | δ = 1 |
|---|---|---|---|---|
| Exact posterior (Table S1b) | 3.000 | 1.702 | 1.274 | 1.000 |
| claude-opus-5-5 | 2.997 [2.993, 3.000] | 1.700 [1.696, 1.703] | 1.274 [1.274, 1.274] | 1.000 [1.000, 1.000] |
| gpt-6.1-sol | 3.000 [3.000, 3.000] | 1.702 [1.702, 1.702] | 1.275 [1.274, 1.277] | 1.000 [1.000, 1.000] |
| claude-haiku-5-5 | 3.908 [3.730, 4.089] | 2.640 [2.494, 2.792] | 2.177 [2.082, 2.276] | 2.914 [2.770, 3.056] |
| gpt-6-luna | 0.788 [0.526, 1.047] | 0.672 [0.572, 0.771] | 0.782 [0.679, 0.886] | 0.124 [0.094, 0.156] |
| four-configuration mean | 2.673 [2.588, 2.758] | 1.679 [1.634, 1.723] | 1.377 [1.343, 1.412] | 1.259 [1.223, 1.296] |

Notes: The over-discount $OD_X(\delta)$ (Section S4.2) is computed before rounding, so it can differ in the last digit from the difference of the printed values.

**Table S3b.** Source weight w2 of a split source (2:1), arm R.

| Configuration | δ = 0 | δ = 0.3 | δ = 0.6 |
|---|---|---|---|
| Exact posterior (Table S1b) | 1.000 | 1.000 | 1.000 |
| claude-opus-5-5 | 0.996 [0.989, 1.000] | 0.997 [0.990, 1.000] | 1.000 [1.000, 1.000] |
| gpt-6.1-sol | 1.000 [1.000, 1.000] | 1.000 [1.000, 1.000] | 1.002 [1.000, 1.007] |
| claude-haiku-5-5 | 1.045 [0.942, 1.148] | 0.787 [0.711, 0.861] | 0.732 [0.661, 0.804] |
| gpt-6-luna | 0.239 [0.095, 0.380] | 0.196 [0.097, 0.293] | 0.163 [0.047, 0.279] |
| four-configuration mean | 0.820 [0.776, 0.863] | 0.745 [0.713, 0.775] | 0.724 [0.691, 0.757] |

Notes: w2 is undefined at δ = 1, where no source splits.

Across arms R, G and DA, the D of Opus and Sol lies between 0.4259 and 0.4361. Table S3c shows arm R against the exact 0.428.

**Table S3c.** Source-weight contrast D per configuration, arm R.

| Configuration | D |
|---|---|
| Exact posterior (Table S1b) | 0.428 |
| claude-opus-5-5 | 0.4306 [0.4283, 0.4352] |
| gpt-6.1-sol | 0.4310 [0.4279, 0.4372] |
| claude-haiku-5-5 | 0.2191 [−0.0684, 0.5081] |
| gpt-6-luna | −0.0431 [−0.1564, 0.0690] |

Notes: The four-configuration mean D is $D_R$ in Table 3 and $D_G$, $D_{CO}$ and $D_{DA}$ in Table S4g.

**Table S3d.** Regression diagnostic (arm R, main panel).

| Configuration | β (SE) | γ (SE) | γ 95% half-width | γ reading |
|---|---|---|---|---|
| claude-opus-5-5 | 0.9986 (0.0012) | 0.0009 (0.0007) | 0.0014 | Exact-reference zone only |
| gpt-6.1-sol | 1.0004 (0.0004) | 0.0001 (0.0001) | 0.0002 | Exact-reference zone only |
| claude-haiku-5-5 | 1.0564 (0.0223) | 0.7477 (0.0353) | 0.0692 | Other |
| gpt-6-luna | 0.2829 (0.0250) | 0.0452 (0.0213) | 0.0417 | Exact-reference zone only |

Notes: Standard errors and γ zones as in Section S2.4. The prior reference, a receiver with no report-dependent term in the regression, has (β, γ) = (0, 0), so the gpt-6-luna row is read with its β of 0.2829. $\varepsilon = 0.001$.

**Table S3e.** Weighted RMS residuals of the fixed projection, in logit units, arm R.

| Configuration | δ = 0 | δ = 0.3 | δ = 0.6 | δ = 1 |
|---|---|---|---|---|
| claude-opus-5-5 | 0.008 | 0.008 | 0.001 | 0.000 |
| gpt-6.1-sol | 3.3e−06 | 2.2e−06 | 0.005 | 3.3e−13 |
| claude-haiku-5-5 | 0.486 | 0.278 | 0.342 | 0.330 |
| gpt-6-luna | 0.806 | 0.306 | 0.487 | 0.067 |
| RedHatAI/Qwen2.5-32B-Instruct-FP8-dynamic | 1.759 | 0.633 | 0.735 | 1.024 |
| RedHatAI/gemma-3-27b-it-FP8-dynamic | 0.896 | 0.714 | 0.664 | 0.097 |

Notes: Cell means, fitted values and residuals of every projection, arms G, CO and DA included, are in results/formal_analysis.json (key projection) in the review archive.

**Table S3f.** Most frequently matched program rule (within 0.001).

| Configuration | Arm | Rule | Matches | Valid |
|---|---|---|---|---|
| claude-opus-5-5 | R | exact posterior | 2,974 | 2,976 |
| claude-opus-5-5 | G | exact posterior | 2,972 | 2,976 |
| claude-opus-5-5 | CO | first child | 2,976 | 2,976 |
| claude-opus-5-5 | DA | exact posterior | 2,970 | 2,976 |
| gpt-6.1-sol | R | exact posterior | 2,975 | 2,976 |
| gpt-6.1-sol | G | exact posterior | 2,976 | 2,976 |
| gpt-6.1-sol | CO | first child | 2,976 | 2,976 |
| gpt-6.1-sol | DA | exact posterior | 2,972 | 2,976 |
| claude-haiku-5-5 | R | exact posterior | 1,114 | 2,976 |
| claude-haiku-5-5 | G | exact posterior | 849 | 2,976 |
| claude-haiku-5-5 | CO | first child | 2,805 | 2,976 |
| claude-haiku-5-5 | DA | exact posterior | 940 | 2,976 |
| gpt-6-luna | R | prior | 1,578 | 2,972 |
| gpt-6-luna | G | prior | 1,576 | 2,973 |
| gpt-6-luna | CO | prior | 1,266 | 2,976 |
| gpt-6-luna | DA | prior | 1,400 | 2,976 |

Notes: Matches are compared among the exact posterior, full count, first child, the prior and root majority. Counts overlap where reference outputs coincide, and the named rule is a numerical reference. The exact posterior is stored under the key bayes in the analysis output.

**Table S3g.** Main panel, arm R (ε = 0.001): percentile interval and Fieller-type set of v and Λ.

| Configuration | v, δ = 0.3 | Fieller, δ = 0.3 | v, δ = 0.6 | Fieller, δ = 0.6 | Λ | Fieller, Λ |
|---|---|---|---|---|---|---|
| claude-opus-5-5 | 0.6304 [0.6299, 0.6314] | [0.6295, 0.6314] | 0.5602 [0.5602, 0.5602] | [0.5602, 0.5602] | 0.0702 [0.0697, 0.0712] | [0.0697, 0.0712] |
| gpt-6.1-sol | 0.6299 [0.6299, 0.6299] | [0.6299, 0.6299] | 0.5599 [0.5591, 0.5603] | [0.5592, 0.5606] | 0.0700 [0.0696, 0.0708] | [0.0696, 0.0708] |
| claude-haiku-5-5 | 0.7704 [0.7500, 0.7907] | [0.7498, 0.7908] | 0.7484 [0.7283, 0.7686] | [0.7283, 0.7689] | 0.0220 [−0.0069, 0.0504] | [−0.0069, 0.0504] |
| gpt-6-luna | 0.7745 [0.6903, 0.8740] | [0.6901, 0.8728] | 0.8271 [0.7299, 0.9437] | [0.7294, 0.9426] | −0.0525 [−0.1936, 0.0839] | [−0.1936, 0.0839] |

Notes: Complete values for arms G, CO and DA are in results/formal_analysis.json in the review archive.

## S3.2 Registered results, scale and complete cases

At the sensitivity scale $\varepsilon = 10^{-6}$ and on complete cases (2,969 worlds), $D_R$ is positive, Holm rejects $D_R = 0$ and the result branch is A+; $T_G$ is not detected, and its 95% interval lies within ±0.0046 and excludes differences larger than 0.0031 in magnitude (Table S3h).

**Table S3h.** Confirmatory family on the sensitivity scale $\varepsilon = 10^{-6}$ and on complete cases.

| Version | Quantity | Estimate | 95% interval | 97.5% interval | p | Decision |
|---|---|---|---|---|---|---|
| $\varepsilon = 10^{-6}$ | $T_G$ | 0.00037 | [−0.0023, 0.0031] | [−0.0027, 0.0035] | 0.7958 | Not detected |
| $\varepsilon = 10^{-6}$ | $D_R$ | 0.28738 | [0.1820, 0.3943] | [0.1682, 0.4099] | 0/10,000 | Positive |
| $\varepsilon = 10^{-6}$ | $D_{diag}$ | 0.17735 | [0.0813, 0.2749] | [0.0669, 0.2910] | 0.0002 | n/a |
| $\varepsilon = 10^{-6}$ | branch | A+ | n/a | n/a | n/a | n/a |
| complete cases (2,969 worlds) | $T_G$ | 0.00035 | [−0.0023, 0.0030] | [−0.0026, 0.0033] | 0.7984 | Not detected |
| complete cases (2,969 worlds) | $D_R$ | 0.25761 | [0.1808, 0.3369] | [0.1713, 0.3477] | 0/10,000 | Positive |
| complete cases (2,969 worlds) | $D_{diag}$ | 0.17597 | [0.0782, 0.2780] | [0.0636, 0.2940] | 0.0006 | n/a |

Notes: The ε = 0.001 family of Table 3 carries the confirmatory reading. p is the empirical p value of the registered counting rule. Where no resample crossed zero the entry reads 0/10,000 (resamples crossing zero), and the smallest positive p of the rule is 0.0002. Both are finite-resample estimates of the tail probability. Decision follows Holm control: Positive, rejected with a positive estimate; Not detected, not rejected. Complete cases (Section S2.2) use ε = 0.001.

All five registered scale checks of Section S2.3 agree between the scales. Every ν gate in every arm passes at both scales. At both, the registered reading is the sign of $D_R$ (Step 1 of Section S2.5), because the interval of the four-configuration mean $\Lambda_R$ includes 0. The weights and diagnostics that move with the scale are those of Haiku and Luna. The γ reading of Haiku is full-count zone only at $\varepsilon = 10^{-6}$ and other at ε = 0.001 (Table S3i).

**Table S3i.** Weights and diagnostics in arm R at $\varepsilon = 10^{-6}$. Pairs are at δ = 0.3 and 0.6.

| Configuration | w3 | w2 | D | ν | Λ | β | γ | γ 95% half-width | γ reading |
|---|---|---|---|---|---|---|---|---|---|
| claude-haiku-5-5 | 2.960, 2.380 | 0.763, 0.727 | 0.3368 | 0.795, 0.766 | 0.0291 | 1.2161 | 1.0236 | 0.1415 | Full-count zone only |
| gpt-6-luna | 0.670, 0.788 | 0.199, 0.161 | −0.0489 | 0.771, 0.831 | −0.0595 | 0.3015 | 0.0327 | 0.0469 | Exact-reference zone only |
| four-configuration mean | 1.758, 1.429 | 0.740, 0.723 | 0.2874 | n/a | 0.0275 | n/a | n/a | n/a | n/a |

Notes: The rows of claude-opus-5-5 and gpt-6.1-sol carry the same values at both scales (Tables S3a–S3d and S3g). For the four-configuration mean, Λ is the mean of the four configurations' Λ values. A panel-level ν is not defined (n/a).

## S3.3 Calibration support, history and X04

**Coverage calibration.** Before any formal request, the frozen inference was run on simulated complete studies of the formal design. Each study draws a new main panel by the production quotas, generates R and G answers of four configurations from specified noisy receivers, and runs the production confirmatory analysis with B = 10,000. The receivers are the exact posterior and full count (each with and without general conservatism), literal-only deduplication, root majority, first child, partial-discount and near-Bayes receivers, and constructed receivers that ignore δ. The logit noise has standard deviation $\sigma_m$, with 30% of its variance shared across configurations within a world and 30% across arms within a configuration. Stress scenarios add extreme answers, fallbacks and heterogeneous σ, and deterministic boundary scenarios at σ = 0 test the degenerate bootstrap. True values are computed exactly over the cell distribution and the distribution of each source's earliest report.

The final round ran 45 scenarios of 20,000 studies each: 42 gating scenarios and three scope-boundary scenarios with very rare switches. The gating set comprises 28 scenarios of the earlier round, rerun at ε = 0.001, and 14 for answer mechanisms seen in the development pilot (Section S6.1), among them rare rule switches and pilot-like panels. A gating scenario met the criteria when the Wilson lower bounds of the 95% coverage of $T_G$ and of $D_R$, and of the simultaneous coverage of the two 97.5% intervals, were at least 0.94, and the Wilson upper bound of the Holm family-wise error on true nulls was at most 0.06. At 20,000 studies, a procedure with exactly nominal coverage passes all 161 checks of the 42 gating scenarios with probability at least 0.9985. The study count was raised from 10,000 on 23 October (Table S6a), after an interim summary of an earlier round and with the thresholds unchanged, because at 10,000 studies Monte Carlo noise could fail a nominally accurate procedure.

Thirty-nine of the 42 gating scenarios, the two pilot-like panels among them, met the criteria and carry the coverage claim. A nominally accurate procedure passes all 151 of their checks with probability at least 0.9986. The other three, A09, C04 and C05, did not meet the criteria and mark the scope boundary. In A09, coverage was 0.935 for $T_G$ and 0.927 for $D_R$ with family-wise error 0.065 on true nulls, and the $T_G$ coverage was 0.864 in C04 and 0.933 in C05. In all three, the four configurations are nearly deterministic, with rare switches to another rule at rates of 0.5–1%. They share one mechanism: resampling within a cell reproduces exactly the switches observed in that cell, so with rare events the variance estimate fluctuates strongly and the percentile tails are misplaced. After these results were seen, and before registration (dates in Section S6.1), A09, C04 and C05 were reclassified as scope-boundary scenarios and the calibration-support screen was added. The inference procedure and the number of worlds stayed as planned.

**Calibration-support screen.** The screen measures how concentrated the inputs of each estimator, fallbacks included, are within narrow windows of each stratum. For $T_G$ the inputs are each configuration's paired difference d at $\delta = 0$ and 1 and the four-configuration mean d of each world; for $D_R$, each configuration's logit_clip($q_R$) at $\delta = 0.3$ and 0.6. In each stratum the screen finds the densest window of fixed width ($10^{-6}$ for d, $10^{-3}$ for the logit) that starts at an observed value. The concentration is the number of worlds in these windows, summed over strata, divided by the number of worlds. A concentration of 0.90 or more marks the input as nearly constant within cells. $T_G$ is inside calibration support when at least two configurations vary and the four-configuration mean varies, and $D_R$ when at least two configurations vary. The screen is an empirical operating condition read from the coverage-calibration studies after their results were seen. When both quantities are inside support, the family is read by the registered rules. Otherwise the family and every reading built on these intervals are descriptive. In the first 20 studies of each scenario, all noisy gating scenarios and both pilot-like panels are inside support in 20 of 20 studies, and A09, C04, C05 and the scope-boundary scenarios outside support in 20 of 20. Concentrations were at most 0.63 for noisy receivers and at least 0.987 per configuration (0.967 for the four-configuration mean) in A09, C04 and C05. Deterministic gating scenarios without switches fall outside support, which is conservative.

**Composition in the formal run.** Both registered quantities are inside support on the formal main panel. On the $T_G$ screen Haiku, Luna and the four-configuration mean vary (concentrations 0.452, 0.562 and 0.292). On the $D_R$ screen Opus, Haiku and Luna vary (0.886, 0.455 and 0.349) and Sol is nearly constant (0.999). Opus lies 0.014 below the 0.90 threshold. With Opus counted as nearly constant, two varying inputs remain and $D_R$ passes the screen. Scenario X04 (deviation D6, below) calibrates the confirmatory reading of $D_R$ at this observed composition of three varying configurations and one nearly constant configuration, a composition beyond the preregistration scenarios. Per-configuration intervals and readings for configurations that the $D_R$ screen marks as nearly constant are descriptive. So are the intervals of the wording control, the r = 0.70 panel and the open-weight replication, which lie beyond the scenarios of the coverage calibration.

**Scenario X04.** At the screen composition of the formal run, scenario X04 gives 95% coverage of 0.9516 for $D_R$, 0.9493 for $T_G$ and 0.9479 for $D_{diag}$ over 20,000 studies, with every true value nonzero. Table S3j gives the coverage and rejection rates. X04 was built after the formal results were seen (deviation D6). It models the two configurations that track the exact posterior as smooth receivers. The rare departures from b that Opus and Sol show in arms R and G enter through the conditional deviation envelopes of Section S3.4 (Table S3k).

**Table S3j.** Scenario X04: coverage and power at the formal screen composition (deviation D6; 20,000 studies, B = 10,000).

| Quantity | True value | 95% coverage [Wilson] | 97.5% coverage | Rejection rate (p ≤ 0.05) | Mean estimate | Monte Carlo SD |
|---|---|---|---|---|---|---|
| $T_G$ | −0.00239 | 0.9493 [0.9462, 0.9523] | 0.9747 | 0.6710 | −0.00239 | 0.00100 |
| $D_R$ | 0.22308 | 0.9516 [0.9485, 0.9544] | 0.9759 | 0.8073 | 0.22319 | 0.07923 |
| $D_{diag}$ | 0.22298 | 0.9479 [0.9448, 0.9509] | 0.9750 | 0.7575 | 0.22296 | 0.08386 |

Notes: The share of studies with the target screen composition is 1.0000. Rejection rates are per-test rates at p ≤ 0.05 before Holm adjustment. The receivers are an exact posterior with logit noise of $5 \times 10^{-4}$, an exact posterior at four decimals and two noisy pilot-like receivers.

## S3.4 Conditional deviation sensitivity

Deviation sensitivity complements coverage calibration for nearly deterministic receivers. A deviation absent from the sample can have positive probability while the percentile interval has zero width, so the analysis reports how far rare deviations could move the true values.

$K = \Sigma_s n_s e_s$ is the expected deviation budget per configuration and arm under the registered within-cell assumptions (Section 4.3): the worlds of a cell deviate at one rate $e_s$, independently of their earliest report values and baseline answers. The main panel reports K = 1 (the main budget) and K = 4 (the smallest positive cell count) for arbitrary answers in [0, 1] and for switches among five rules (exact posterior, full count, first child, prior, root majority). A rule chosen within a cell is independent of the earliest report values and baseline answers but may differ between cells. The worst-case shifts of the true values are computed exactly under these assumptions (stability is defined in the notes to Table S3k). In the formal data, Opus and Sol lie more than 0.001 from b in 0 to 4 of 2,976 worlds in each of arms R and G, the two arms that enter $D_R$ and $T_G$ (Table S3f). The largest observed count equals the K = 4 budget. The envelope is conditional on the formal worlds and the baseline answers, and all registered decisions stand as reported.

$D_R$ stays positive in three of its four envelopes, where the widened interval excludes zero (rule switches at both budgets and arbitrary answers at K = 1). Under arbitrary answers at K = 4 the observed interval excludes zero and the widened interval, [−0.267, 0.593], crosses it. The widened $T_G$ interval stays within the registered precision bound of ±0.0046 under rule switches at K = 1 and extends beyond it in the other three envelopes.

The computation covers $T_G$, $D_R$, $\Lambda_R$, the four-configuration mean $w3/w3_B$ and w2, and the mismatch at the middle levels. Linear quantities are solved as fractional knapsack problems. D and ν are solved at endpoints and stationary points, with the budget split between the two δ levels. For four exact receivers at K = 1, the reference shifts are 0.00253 for $T_G$, 0.0177 for $\Lambda_R$ and 0.117 for $D_R$ under arbitrary answers, and 0.00058, 0.0023 and 0.0157 under rule switches.

**Table S3k.** Deviation sensitivity of the conclusions that hold on the observed intervals.

| Conclusion | Observed 95% interval | arbitrary answers, K = 1 | arbitrary answers, K = 4 | rule switches, K = 1 | rule switches, K = 4 |
|---|---|---|---|---|---|
| $D_R \neq 0$ | [0.1824, 0.3369] | 0.1123 / 0.0641; holds and stable | 0.4490 / 0.2563; holds, not stable | 0.0148 / 0.0096; holds and stable | 0.0594 / 0.0383; holds and stable |
| $T_G$ interval within the registered precision bound ±0.0046 | [−0.0023, 0.0031] | 0.0025 / 0.0025; holds, not stable | 0.0101 / 0.0101; holds, not stable | 0.0006 / 0.0006; holds and stable | 0.0023 / 0.0025; holds, not stable |
| mean $w3/w3_B$ in [0.75, 1.25], δ = 0.3 | [0.9602, 1.0125] | 0.0242 / 0.0150; holds and stable | 0.0967 / 0.0601; holds and stable | 0.0073 / 0.0050; holds and stable | 0.0291 / 0.0201; holds and stable |
| mean $w3/w3_B$ in [0.75, 1.25], δ = 0.6 | [1.0540, 1.1086] | 0.0203 / 0.0139; holds and stable | 0.0814 / 0.0557; holds and stable | 0.0053 / 0.0054; holds and stable | 0.0211 / 0.0214; holds and stable |

Notes: Each cell gives the worst-case downward / upward shift of the true value, then the status. Holds: the observed 95% interval satisfies the conclusion. Stable: it still does after widening by the worst-case shift in the threatening direction. Five further registered conditions have observed intervals outside their acceptance regions and carry no envelope: $\Lambda_R$ within ±0.0174 (Table 3), and, at δ = 0.3 and at δ = 0.6, mean w2 in [0.75, 1.25] and the mismatch upper limit (Section S3.5). The shifts of every quantity are in results/formal_analysis.json, key deviation_sensitivity.

### S3.5 Closeness conditions and diagnostic interpretation

The closeness conditions of Section S2.5 are set on the four-configuration mean. Its w2 intervals extend below 0.75 at $\delta = 0.3$ and 0.6 in every arm ([0.713, 0.775] and [0.691, 0.757] in arm R; Table S3b). Its raw risk, 0.0165 and 0.0162, exceeds the limits of 0.0023 and 0.0052 (Table S4b). The four-configuration mean therefore lies outside the registered conditions for closeness wording in every arm, and closeness is stated per configuration. Individually, Opus and Sol lie within 0.001 of the exact posterior in 2,974 and 2,975 of 2,976 arm-R worlds (Table S3f). The counts at each δ are in Table S4a.

Four frozen diagnostics discriminated the specified receivers well in finite samples: the ν gate, the direction of Λ, the γ reading and the compliance difference. The noiseless exact receiver was classified correctly on these four and on the saturation diagnostic. The simulation ran 17 receiver sets of 200 studies each ($B = 1{,}000$, $\varepsilon = 0.001$): 12 at $\sigma = 0.6$ and 5 added after the pilot.

The equivalence reading of Λ is strict: against the registered bound of ±0.0174, the 95% half-width of the four-configuration mean Λ at $\sigma = 0.6$ is about 0.014–0.020. Equivalence was detected in 27% of studies for full count, 3% for full count on a 0.05 grid and 0% for the other receivers that ignore δ. The saturation diagnostic detected positive values at rates 0.81 for Bayes, 0.73 for near-Bayes, 0.36 for Bayes × 0.5 and 0.20 for prior answers varying with δ. For the mixture receiver modelled on one pilot configuration, whose true γ is 1.115, 61% of studies read full-count zone only and 39% read other, the two readings that Haiku takes at the two truncation scales (Section S3.2).

## S4. Rule agreement and paired gain separate

This section holds the evidence for the article's main finding (Sections 6.1–6.5). Table S4a sets agreement beside paired gain at every dependence level. In each cell, the reference gain under faithful implementation is the raw risk of Table S4b minus the realised-world target risk of Section S1.2 (0.0440754, 0.0229553, 0.0122546 and 0 at $\delta = 0$, 0.3, 0.6 and 1). Among the four closed-weight configurations, wherever the rule or the raw baseline carries risk, the point estimate of $\Delta_{CO}$ (Tables S4a and S4c) has the sign of that reference gain (post hoc descriptive; Section 8.2). The audit runs both arms because the instructed arm is measured rather than forecast. For Luna at $\delta = 0.3$, 0.6 and 1 the paired gain lies far below the reference gain and its interval includes 0 (Table S4a). Two open-weight cells have point estimates of the opposite sign to the reference gain (Section S5.3; Table S5d; post hoc descriptive).

### S4.1 Condition-aligned agreement and paired gain

Counts, including those over the worlds where the rule's output differs from the prior, are post hoc descriptive tallies of the existing final logs in the 744 quota worlds per level. Gains are natural-distribution means.

**Table S4a.** Condition-aligned counts and paired gain $\Delta_{CO}$ at each dependence level (main panel, template P0-draft-2).

| Configuration | δ | R valid | R within 0.001 of b | CO within 0.001 of first child | Worlds with first child ≠ prior | CO within 0.001 of first child in those worlds | $\Delta_{CO}$ [95% interval] |
|---|---|---|---|---|---|---|---|
| claude-opus-5-5 | 0 | 744 | 743 | 744 | 415 | 415 | −0.0440 [−0.0481, −0.0401] |
| claude-opus-5-5 | 0.3 | 744 | 743 | 744 | 406 | 406 | −0.0228 [−0.0254, −0.0204] |
| claude-opus-5-5 | 0.6 | 744 | 744 | 744 | 404 | 404 | −0.0123 [−0.0139, −0.0106] |
| claude-opus-5-5 | 1 | 744 | 744 | 744 | 414 | 414 | 7.7e−10 [5.2e−10, 1.0e−09] |
| gpt-6.1-sol | 0 | 744 | 744 | 744 | 415 | 415 | −0.0441 [−0.0481, −0.0401] |
| gpt-6.1-sol | 0.3 | 744 | 744 | 744 | 406 | 406 | −0.0230 [−0.0255, −0.0205] |
| gpt-6.1-sol | 0.6 | 744 | 743 | 744 | 404 | 404 | −0.0123 [−0.0139, −0.0106] |
| gpt-6.1-sol | 1 | 744 | 744 | 744 | 414 | 414 | 4.0e−26 [−4.8e−27, 9.0e−26] |
| claude-haiku-5-5 | 0 | 744 | 367 | 727 | 415 | 401 | −0.0295 [−0.0342, −0.0250] |
| claude-haiku-5-5 | 0.3 | 744 | 157 | 647 | 406 | 312 | −0.0066 [−0.0108, −0.0023] |
| claude-haiku-5-5 | 0.6 | 744 | 172 | 702 | 404 | 362 | 0.0095 [0.0065, 0.0127] |
| claude-haiku-5-5 | 1 | 744 | 418 | 729 | 414 | 400 | 0.0302 [0.0260, 0.0348] |
| gpt-6-luna | 0 | 744 | 132 | 165 | 415 | 1 | 0.0140 [0.0064, 0.0219] |
| gpt-6-luna | 0.3 | 742 | 89 | 149 | 406 | 0 | 0.0022 [−0.0031, 0.0075] |
| gpt-6-luna | 0.6 | 742 | 85 | 137 | 404 | 1 | 0.0047 [−0.0001, 0.0097] |
| gpt-6-luna | 1 | 744 | 293 | 235 | 414 | 2 | 0.0030 [−0.0001, 0.0060] |

Notes: Counts are of answers within 0.001 of the reference (Section S2.3). The reference of arm R is the exact posterior b, and that of arm CO is the first-child output. At δ = 1 they coincide (target risk 0, Table S1b). Per arm and level, 744 answers were requested, and CO valid is 744 in every row. R valid is 744 except for the gpt-6-luna arm-R refusals under deviation D1 (two at δ = 0.3 and two at δ = 0.6; Table S2a), and all other requests returned status ok. Gains and intervals are those of Table S4c.

## S4.2 Risk terms, sign of the gain and prespecified secondary quantities

Table S4b gives the natural-distribution risk of arms R and CO and Table S4c the paired gain of arms G, CO and DA. Tables S4d–S4g report the standardised gains of Section S2.3 and the prespecified secondary quantities of Section S2.4 on the main-panel worlds and the resamples of Section S2.5, with statuses as in Conventions. Holm results within the secondary family are flags. The over-discount $OD_X(\delta) = w3_B(\delta) - w3_X(\delta)$, in units of one report, is positive when arm X discounts agreeing sources more than the exact posterior does. It follows from the source weights w3 of Section S3.1 and the exact $w3_B$ of Table S1b. Its values with intervals are in results/formal_analysis.json ($OD_X$ per configuration and panel_OD_X for the four-configuration mean, under blocks.v3.5.secondary). Standardised mismatch and Δ per configuration, Brier scores and signed logit distortion are in results/formal_analysis.json (mm_std_* and Delta_std_* under blocks.v3.5.secondary, and scores).

**Table S4b.** Natural-distribution risk of arms R and CO.

| Configuration | Arm | δ = 0 | δ = 0.3 | δ = 0.6 | δ = 1 |
|---|---|---|---|---|---|
| claude-opus-5-5 | R | 3.2e−05 [−1.1e−06, 0.0001] | 0.0001 [−5.2e−06, 0.0004] | 2.1e−08 [1.8e−08, 2.3e−08] | 1.1e−09 [8.9e−10, 1.4e−09] |
| gpt-6.1-sol | R | 2.8e−14 [2.0e−14, 3.7e−14] | 5.0e−12 [7.5e−14, 1.2e−11] | 2.3e−06 [−7.0e−08, 6.9e−06] | 5.1e−26 [1.5e−26, 9.6e−26] |
| claude-haiku-5-5 | R | 0.0139 [0.0114, 0.0163] | 0.0185 [0.0155, 0.0220] | 0.0221 [0.0198, 0.0247] | 0.0312 [0.0271, 0.0357] |
| gpt-6-luna | R | 0.0789 [0.0728, 0.0851] | 0.0473 [0.0432, 0.0516] | 0.0426 [0.0387, 0.0467] | 0.0293 [0.0277, 0.0311] |
| four-configuration mean | R | 0.0232 [0.0215, 0.0249] | 0.0165 [0.0152, 0.0179] | 0.0162 [0.0150, 0.0174] | 0.0151 [0.0140, 0.0163] |
| claude-opus-5-5 | CO | 0.0441 [0.0401, 0.0481] | 0.0230 [0.0205, 0.0255] | 0.0123 [0.0106, 0.0139] | 3.6e−10 [3.6e−10, 3.7e−10] |
| gpt-6.1-sol | CO | 0.0441 [0.0401, 0.0481] | 0.0230 [0.0205, 0.0255] | 0.0123 [0.0106, 0.0139] | 1.1e−26 [−1.3e−28, 3.4e−26] |
| claude-haiku-5-5 | CO | 0.0434 [0.0394, 0.0474] | 0.0251 [0.0221, 0.0282] | 0.0126 [0.0108, 0.0144] | 0.0010 [0.0002, 0.0020] |
| gpt-6-luna | CO | 0.0648 [0.0602, 0.0695] | 0.0451 [0.0415, 0.0488] | 0.0379 [0.0349, 0.0410] | 0.0263 [0.0238, 0.0290] |
| four-configuration mean | CO | 0.0491 [0.0459, 0.0525] | 0.0290 [0.0269, 0.0312] | 0.0188 [0.0173, 0.0203] | 0.0068 [0.0062, 0.0075] |

Notes: Rescaled bootstrap lower limits can be negative for non-negative scores. Complete values for all arms are in results/formal_analysis.json in the review archive.

**Table S4c.** $\Delta_X(\delta)$ for X in {G, CO, DA}. Positive values mean smaller mismatch than in arm R.

| Configuration | Arm | δ = 0 | δ = 0.3 | δ = 0.6 | δ = 1 |
|---|---|---|---|---|---|
| claude-opus-5-5 | G | 3.2e−05 [−1.1e−06, 0.0001] | 0.0001 [−5.2e−06, 0.0004] | −3.6e−07 [−1.1e−06, 1.7e−08] | 2.0e−10 [−1.4e−10, 5.4e−10] |
| gpt-6.1-sol | G | 1.1e−14 [7.8e−19, 2.2e−14] | −3.5e−11 [−1.0e−10, 7.9e−12] | 2.3e−06 [−7.0e−08, 6.9e−06] | −1.3e−28 [−5.9e−26, 5.8e−26] |
| claude-haiku-5-5 | G | −0.0012 [−0.0051, 0.0024] | −0.0044 [−0.0081, −0.0004] | −0.0034 [−0.0064, −0.0003] | 0.0019 [−0.0032, 0.0071] |
| gpt-6-luna | G | 0.0065 [−0.0019, 0.0150] | 0.0091 [0.0036, 0.0145] | 0.0071 [0.0015, 0.0127] | 0.0050 [0.0027, 0.0074] |
| four-configuration mean | G | 0.0013 [−0.0009, 0.0036] | 0.0012 [−0.0004, 0.0029] | 0.0009 [−0.0006, 0.0025] | 0.0017 [0.0003, 0.0031] |
| claude-opus-5-5 | CO | −0.0440 [−0.0481, −0.0401] | −0.0228 [−0.0254, −0.0204] | −0.0123 [−0.0139, −0.0106] | 7.7e−10 [5.2e−10, 1.0e−09] |
| gpt-6.1-sol | CO | −0.0441 [−0.0481, −0.0401] | −0.0230 [−0.0255, −0.0205] | −0.0123 [−0.0139, −0.0106] | 4.0e−26 [−4.8e−27, 9.0e−26] |
| claude-haiku-5-5 | CO | −0.0295 [−0.0342, −0.0250] | −0.0066 [−0.0108, −0.0023] | 0.0095 [0.0065, 0.0127] | 0.0302 [0.0260, 0.0348] |
| gpt-6-luna | CO | 0.0140 [0.0064, 0.0219] | 0.0022 [−0.0031, 0.0075] | 0.0047 [−0.0001, 0.0097] | 0.0030 [−0.0001, 0.0060] |
| four-configuration mean | CO | −0.0259 [−0.0297, −0.0223] | −0.0126 [−0.0150, −0.0101] | −0.0026 [−0.0044, −0.0007] | 0.0083 [0.0070, 0.0097] |
| claude-opus-5-5 | DA | 8.7e−06 [−0.0001, 0.0001] | 9.6e−10 [−0.0004, 0.0004] | −9.9e−06 [−3.1e−05, 4.5e−07] | 5.2e−10 [2.5e−10, 8.1e−10] |
| gpt-6.1-sol | DA | 2.8e−15 [−1.0e−14, 1.5e−14] | −3.7e−05 [−0.0001, 1.4e−06] | −0.0002 [−0.0005, 1.0e−05] | 3.1e−26 [−1.6e−26, 8.3e−26] |
| claude-haiku-5-5 | DA | −0.0001 [−0.0035, 0.0031] | −0.0001 [−0.0047, 0.0045] | 0.0004 [−0.0026, 0.0035] | 0.0027 [−0.0027, 0.0080] |
| gpt-6-luna | DA | 0.0006 [−0.0083, 0.0095] | 0.0046 [−0.0011, 0.0102] | 0.0066 [0.0015, 0.0118] | 0.0032 [0.0009, 0.0055] |
| four-configuration mean | DA | 0.0001 [−0.0022, 0.0024] | 0.0011 [−0.0006, 0.0029] | 0.0017 [0.0002, 0.0032] | 0.0015 [−4.5e−05, 0.0030] |

**Table S4d.** Four-configuration mean $\Delta_X(\delta)$ standardised on common difficulty support.

| Arm | δ = 0 | δ = 0.3 | δ = 0.6 | δ = 1 |
|---|---|---|---|---|
| G | −0.0002 [−0.0025, 0.0022] | 0.0013 [−0.0003, 0.0029] | 0.0011 [−0.0005, 0.0026] | 0.0022 [0.0010, 0.0036] |
| CO | −0.0249 [−0.0293, −0.0206] | −0.0119 [−0.0143, −0.0095] | −0.0023 [−0.0041, −0.0005] | 0.0052 [0.0038, 0.0067] |
| DA | −0.0013 [−0.0038, 0.0013] | 0.0009 [−0.0008, 0.0028] | 0.0017 [0.0002, 0.0032] | 0.0019 [0.0004, 0.0033] |

**Table S4e.** Compliance error and compliance difference in arm CO.

| Configuration | Arm | δ = 0 | δ = 0.3 | δ = 0.6 | δ = 1 |
|---|---|---|---|---|---|
| **Compliance error $E[(q_{CO} - q_{FC})^2]$** | | | | | |
| claude-opus-5-5 | CO | 5.1e−10 [4.0e−10, 6.3e−10] | 4.8e−10 [3.8e−10, 5.9e−10] | 4.3e−10 [3.6e−10, 5.2e−10] | 3.6e−10 [3.6e−10, 3.7e−10] |
| gpt-6.1-sol | CO | 1.5e−26 [−5.1e−28, 3.0e−26] | 2.0e−10 [−4.9e−12, 6.0e−10] | 1.8e−25 [1.1e−25, 2.6e−25] | 1.1e−26 [−1.3e−28, 3.4e−26] |
| claude-haiku-5-5 | CO | 0.00053 [0.00018, 0.00095] | 0.00284 [0.00197, 0.00384] | 0.00051 [0.00010, 0.00121] | 0.00099 [0.00018, 0.00198] |
| gpt-6-luna | CO | 0.03413 [0.03085, 0.03746] | 0.03270 [0.02959, 0.03588] | 0.02970 [0.02692, 0.03259] | 0.02629 [0.02379, 0.02898] |
| four-configuration mean | CO | 0.00866 [0.00784, 0.00951] | 0.00889 [0.00808, 0.00972] | 0.00755 [0.00684, 0.00829] | 0.00682 [0.00615, 0.00754] |
| **Compliance difference $E[(q_R - q_{FC})^2] - E[(q_{CO} - q_{FC})^2]$** | | | | | |
| four-configuration mean | CO | 0.03834 [0.03515, 0.04158] | 0.02352 [0.02121, 0.02588] | 0.01748 [0.01570, 0.01930] | 0.00829 [0.00696, 0.00966] |

Notes: Complete compliance-difference values for all configurations are in results/formal_analysis.json in the review archive.

$T_M(\delta)$ is the mismatch of CO minus that of DA.

**Table S4f.** $T_{CO}$ and $T_M$.

| Configuration | $T_{CO}$ | $T_M$, δ = 0.3 | $T_M$, δ = 0.6 |
|---|---|---|---|
| claude-opus-5-5 | 0.0440 [0.0401, 0.0481] | 0.0228 [0.0204, 0.0254] | 0.0122 [0.0106, 0.0139] |
| gpt-6.1-sol | 0.0441 [0.0401, 0.0481] | 0.0229 [0.0204, 0.0255] | 0.0121 [0.0104, 0.0137] |
| claude-haiku-5-5 | 0.0597 [0.0535, 0.0662] | 0.0065 [0.0021, 0.0109] | −0.0091 [−0.0120, −0.0063] |
| gpt-6-luna | −0.0111 [−0.0196, −0.0027] | 0.0024 [−0.0029, 0.0075] | 0.0019 [−0.0023, 0.0062] |
| four-configuration mean | 0.0342 [0.0303, 0.0382] | 0.0137 [0.0111, 0.0163] | 0.0043 [0.0026, 0.0061] |

**Table S4g.** Four-configuration mean secondary quantities with p values.

| Quantity | Estimate | 95% interval | p | Holm flag |
|---|---|---|---|---|
| $D_G$ | 0.2758 | [0.1945, 0.3579] | 0/10,000 | flag |
| $\Lambda_G$ | 0.0411 | [0.0160, 0.0661] | 0.0022 | n/a |
| $D_{CO}$ | −0.0412 | [−0.1481, 0.0644] | 0.4360 | no flag |
| $\Lambda_{CO}$ | −0.0322 | [−0.1024, 0.0338] | 0.3260 | n/a |
| $D_{DA}$ | 0.2409 | [0.1671, 0.3162] | 0/10,000 | flag |
| $\Lambda_{DA}$ | 0.0214 | [−0.0101, 0.0499] | 0.1750 | n/a |
| $\Lambda_G - \Lambda_R$ | 0.0137 | [−0.0270, 0.0571] | 0.5274 | no flag |
| $\Lambda_{CO} - \Lambda_R$ | −0.0596 | [−0.1357, 0.0115] | 0.0992 | no flag |
| $\Lambda_{DA} - \Lambda_R$ | −0.0060 | [−0.0513, 0.0378] | 0.7754 | no flag |

Notes: p and the 0/10,000 entries as in Table S3h. The Holm column marks flags within the secondary family, and n/a marks a quantity outside that family. The Λ rows give the four-configuration mean ratio change per arm and its difference from arm R. $D_R$ and $\Lambda_R$ are in Table 3 of the article. $T_{CO}$, of the same family, is in Table S4f, and Table 3 gives its p value and Holm flag.

### S4.3 Decision-scale consequences

**Table S4h.** 0/1 decision regret in arms R and CO (point estimates).

| Configuration | Arm | δ = 0 | δ = 0.3 | δ = 0.6 | δ = 1 |
|---|---|---|---|---|---|
| claude-opus-5-5 | R | 3.9e−18 | 0.0004 | 0.0000 | 0.0000 |
| gpt-6.1-sol | R | 3.9e−18 | 0.0000 | 0.0000 | 0.0000 |
| claude-haiku-5-5 | R | 0.0325 | 0.0293 | 0.0309 | 0.0200 |
| gpt-6-luna | R | 0.1320 | 0.0924 | 0.0801 | 0.0630 |
| four-configuration mean | R | 0.0411 | 0.0305 | 0.0277 | 0.0207 |
| claude-opus-5-5 | CO | 0.0713 | 0.0541 | 0.0281 | 0.0000 |
| gpt-6.1-sol | CO | 0.0713 | 0.0541 | 0.0281 | 0.0000 |
| claude-haiku-5-5 | CO | 0.0712 | 0.0555 | 0.0295 | 0.0013 |
| gpt-6-luna | CO | 0.1104 | 0.0892 | 0.0763 | 0.0476 |
| four-configuration mean | CO | 0.0810 | 0.0632 | 0.0405 | 0.0122 |

Notes: Regret takes the expectation over Y given x analytically.

## S5. Controls and receiver–prompt conditions

This section tests the reach of Sections 5 and 6. S5.1 separates new child reports from copies and paraphrases. S5.2 carries the raw weighting result to a second reliability level and two open-weight configurations. S5.3 reproduces the separation for the open-weight configurations in all four arms. It gives the two cells where the paired gain $\Delta_{CO}$ has the opposite sign to the reference gain. S5.4–S5.5 show how far the raw baselines move when the prompt configuration and the run period change together. The wording control, the second reliability level and the open-weight configurations are descriptive, and conditions A and B are exploratory (Sections 6.6 and 7).

### S5.1 New reports, copies and paraphrases

The wording control (Section S1.3) covers 240 parent worlds at δ = 0.3 and 0.6 in arms R and G. Section 7.1 gives the reading of Table S5a and Fig. S1a, b. Entries are the Horvitz–Thompson estimates of Section S2.4, with 95% intervals from resampling worlds within each δ × π × L layer (descriptive; Section S5).

**Table S5a.** Wording control.

| Configuration | Arm | δ | New minus copy | Ratio to exact | Copy minus paraphrase |
|---|---|---|---|---|---|
| claude-opus-5-5 | R | 0.3 | 0.5084 [0.4409, 0.5775] | 1.00 | 0.0006 [0.0002, 0.0010] |
| claude-opus-5-5 | R | 0.6 | 0.2159 [0.1606, 0.2754] | 1.00 | −0.0001 [−0.0006, 0.0005] |
| claude-opus-5-5 | G | 0.3 | 0.5082 [0.4409, 0.5774] | 1.00 | 4.1e−06 [−0.0004, 0.0004] |
| claude-opus-5-5 | G | 0.6 | 0.2155 [0.1602, 0.2750] | 1.00 | 0.0005 [−4.7e−05, 0.0011] |
| gpt-6.1-sol | R | 0.3 | 0.5081 [0.4408, 0.5771] | 1.00 | −1.7e−06 [−1.1e−05, 6.3e−06] |
| gpt-6.1-sol | R | 0.6 | 0.2151 [0.1597, 0.2747] | 1.00 | −2.7e−07 [−1.7e−06, 1.1e−06] |
| gpt-6.1-sol | G | 0.3 | 0.5081 [0.4408, 0.5771] | 1.00 | 2.0e−06 [−1.1e−05, 1.4e−05] |
| gpt-6.1-sol | G | 0.6 | 0.2151 [0.1597, 0.2747] | 1.00 | −1.3e−06 [−4.3e−06, 8.6e−07] |
| claude-haiku-5-5 | R | 0.3 | 0.7889 [0.5006, 1.1031] | 1.55 | −0.1367 [−0.3856, 0.0976] |
| claude-haiku-5-5 | R | 0.6 | 0.6024 [0.3010, 0.9277] | 2.80 | 0.0462 [−0.1561, 0.2487] |
| claude-haiku-5-5 | G | 0.3 | 0.6116 [0.3473, 0.8881] | 1.20 | −0.1403 [−0.3908, 0.1028] |
| claude-haiku-5-5 | G | 0.6 | 0.5584 [0.2981, 0.8029] | 2.60 | 0.1010 [−0.1641, 0.3615] |
| gpt-6-luna | R | 0.3 | 0.2599 [−0.0965, 0.6253] | 0.51 | −0.0092 [−0.2713, 0.2678] |
| gpt-6-luna | R | 0.6 | −0.0815 [−0.4900, 0.3175] | −0.38 | −0.2517 [−0.6308, 0.1495] |
| gpt-6-luna | G | 0.3 | −0.0479 [−0.3338, 0.2321] | −0.09 | 0.2940 [0.0273, 0.5675] |
| gpt-6-luna | G | 0.6 | 0.1609 [−0.2216, 0.5513] | 0.75 | 0.0164 [−0.3167, 0.3476] |

Notes: The exact references for new minus copy and for copy minus paraphrase are 0.5081 and 0.0000 at δ = 0.3, and 0.2151 and 0.0000 at δ = 0.6. They are means over the sampled worlds, estimated with the same weights as the configuration entries. The ratio column gives point values: the configuration estimate divided by the exact new-minus-copy mean.

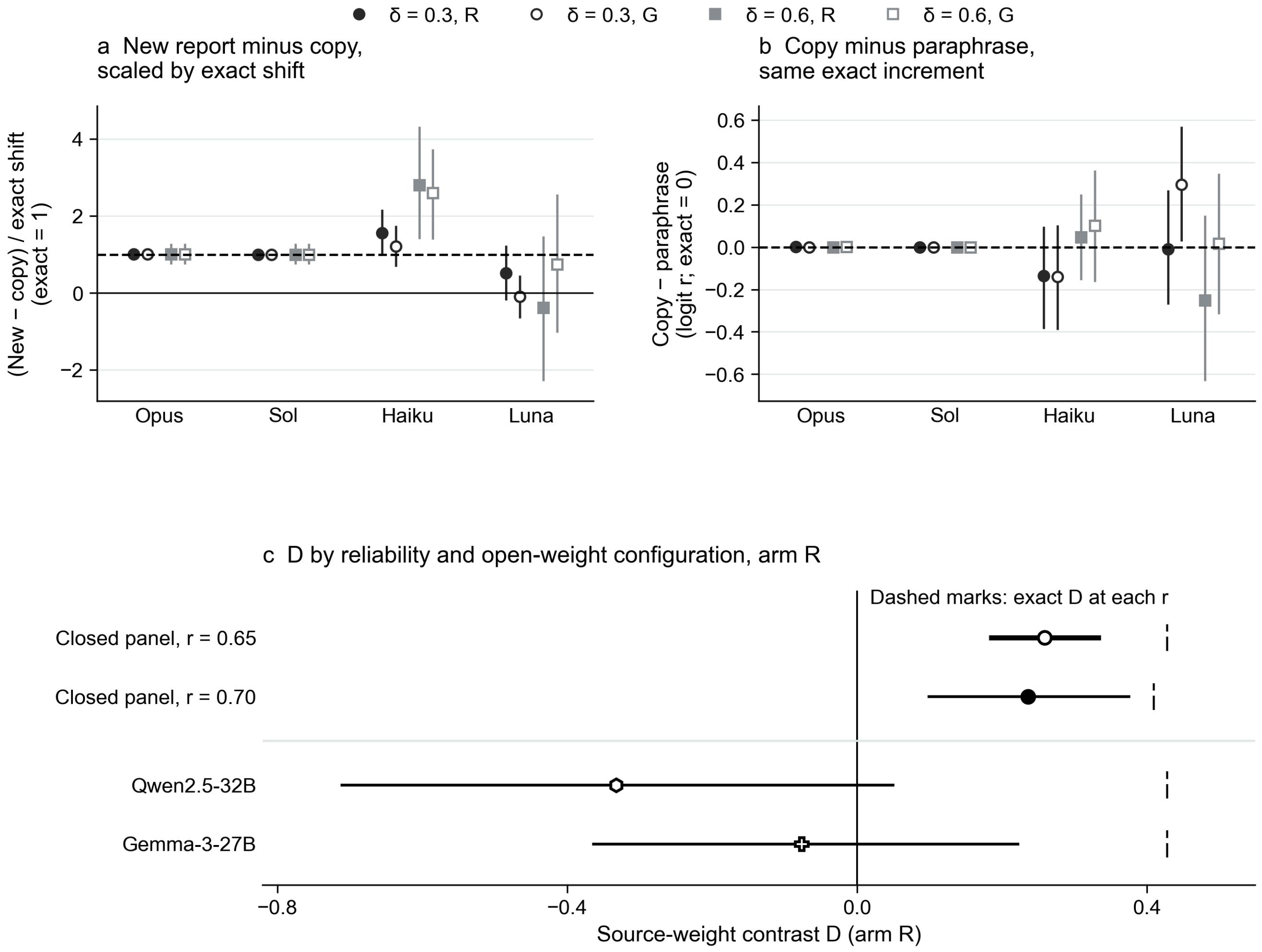

**Fig. S1. Evidence-type controls and scope of the weighting measurements.** (a) Response to a new child report in a fixed slot, relative to a verbatim copy and divided by the exact shift (reference line at 1; the exact shifts are in Table S5a). Filled markers are arm R and open markers arm G. Circles are δ = 0.3 and squares δ = 0.6. (b) Copy minus paraphrase in units of logit r (exact reference 0). In panels a and b, each δ has 120 parent worlds (240 in the panel). Descriptive 95% bootstrap percentile intervals come from resampling worlds within δ × π × L layers. Panel a rescales these intervals by the exact shift on the sampled worlds. (c) Source-weight contrast D in arm R, with a short vertical line at the exact D for each row's reliability. The rows are the main panel (r = 0.65; exact 0.428; confirmatory family with $T_G$; Table 3 of the article), the r = 0.70 panel (768 worlds, arms R and G; exact 0.410; descriptive) and the open-weight configurations Qwen and Gemma, which ran under template P0-draft-3 (deviations D2 and D3; descriptive). Estimates and intervals are in Table 3 and Table S5c.

## S5.2 Second reliability level and raw open-weight results

Tables S5b and S5c are descriptive (Section 7.2). Fig. S1c shows the D values together, and Table S3e gives the projection residuals of the open-weight configurations.

**Table S5b.** v gate in arm R and point estimates of the source weights in the r = 0.70 panel (closed-weight configurations) and in the main panel at r = 0.65 (open-weight configurations).

| Configuration | v gate | w3, δ = 0.3 and 0.6 | w2, δ = 0.3 and 0.6 | w3, δ = 0 | w2, δ = 0 | w3, δ = 1 |
|---|---|---|---|---|---|---|
| claude-opus-5-5 (r = 0.70) | yes | 1.675, 1.264 | 1.000, 0.995 | 2.999 | 1.001 | 1.000 |
| gpt-6.1-sol (r = 0.70) | yes | 1.675, 1.264 | 1.000, 1.002 | 3.000 | 1.000 | 1.000 |
| claude-haiku-5-5 (r = 0.70) | yes | 2.190, 1.932 | 0.702, 0.726 | 3.143 | 0.916 | 2.242 |
| gpt-6-luna (r = 0.70) | no | 0.580, 0.604 | 0.346, 0.179 | 0.623 | 0.115 | 0.167 |
| RedHatAI/Qwen2.5-32B-Instruct-FP8-dynamic (r = 0.65) | yes | 2.985, 2.870 | 0.632, 0.496 | 4.062 | 1.167 | 3.377 |
| RedHatAI/gemma-3-27b-it-FP8-dynamic (r = 0.65) | yes | 3.228, 3.188 | 0.525, 0.495 | 4.541 | 0.822 | 5.512 |

Notes: Descriptive point estimates per configuration.

**Table S5c.** Second-reliability panel and open-weight replication quantities (descriptive).

| Block | Quantity | Estimate | 95% interval |
|---|---|---|---|
| r = 0.70 panel | $T_G$ | −0.00331 | [−0.0092, 0.0024] |
| r = 0.70 panel | $D_R$ | 0.23627 | [0.0975, 0.3776] |
| r = 0.70 panel | $D_{diag}$ | 0.27914 | [0.1826, 0.3713] |
| RedHatAI/Qwen2.5-32B-Instruct-FP8-dynamic | $T_G$ | 0.03447 | [0.0249, 0.0442] |
| RedHatAI/Qwen2.5-32B-Instruct-FP8-dynamic | D | −0.33239 | [−0.7126, 0.0518] |
| RedHatAI/Qwen2.5-32B-Instruct-FP8-dynamic | $D_{diag}$ | −0.01417 | [−0.3417, 0.3274] |
| RedHatAI/gemma-3-27b-it-FP8-dynamic | $T_G$ | −0.07967 | [−0.0921, −0.0671] |
| RedHatAI/gemma-3-27b-it-FP8-dynamic | D | −0.07593 | [−0.3652, 0.2240] |
| RedHatAI/gemma-3-27b-it-FP8-dynamic | $D_{diag}$ | 0.08550 | [−0.1700, 0.3556] |

Notes: The r = 0.70 rows give four-configuration means. The open-weight rows give per-configuration values.

## S5.3 Open-weight paired-gain reproduction

In Table S5d the response to the instruction gives the paired gain $\Delta_{CO}$ the opposite sign to the reference gain for Qwen at δ = 0 and Gemma at δ = 0.3 (Section 8.2; Section S4 states why the audit runs both arms). Both configurations answered the 2,976 worlds of the main panel, so the realised-world target risk of Section S1.2 applies unchanged. The reference gain of each cell is its raw risk minus the target risk, as in Section S4. Section 6.6 gives the reading of this post hoc descriptive table (deviation D8). The table scores the two configurations as deployed: quantised snapshots served by vLLM under template P0-draft-3, with its revised answer sentence and a 256-visible-token cap (Sections S6.2.2 and S6.4).

**Table S5d.** Open-weight configurations in all four arms of the main panel (post hoc, descriptive, point estimates).

| | | Risk | | | | $\Delta_X$ | | | | | |
|---|---|---|---|---|---|---|---|---|---|---|---|
| **Configuration** | **Arm** | **δ = 0** | **δ = 0.3** | **δ = 0.6** | **δ = 1** | **δ = 0** | **δ = 0.3** | **δ = 0.6** | **δ = 1** | **Most frequently matched rule, matches** | **Within 0.001 of first child** |
| RedHatAI/Qwen2.5-32B-Instruct-FP8-dynamic | R | 0.0358 | 0.0284 | 0.0485 | 0.0967 | n/a | n/a | n/a | n/a | prior, 156 | 78 |
| RedHatAI/Qwen2.5-32B-Instruct-FP8-dynamic | G | 0.0357 | 0.0244 | 0.0405 | 0.0622 | 2.2e−05 | 0.0040 | 0.0080 | 0.0345 | full count, 153 | 91 |
| RedHatAI/Qwen2.5-32B-Instruct-FP8-dynamic | CO | 0.0328 | 0.0237 | 0.0186 | 0.0189 | 0.0030 | 0.0047 | 0.0299 | 0.0778 | prior, 313 | 169 |
| RedHatAI/Qwen2.5-32B-Instruct-FP8-dynamic | DA | 0.0378 | 0.0230 | 0.0393 | 0.0864 | −0.0020 | 0.0054 | 0.0091 | 0.0103 | prior, 185 | 86 |
| RedHatAI/gemma-3-27b-it-FP8-dynamic | R | 0.0246 | 0.0290 | 0.0420 | 0.0840 | n/a | n/a | n/a | n/a | exact posterior, 14 | 6 |
| RedHatAI/gemma-3-27b-it-FP8-dynamic | G | 0.0266 | 0.0528 | 0.0823 | 0.1656 | −0.0019 | −0.0238 | −0.0404 | −0.0816 | full count, 11 | 6 |
| RedHatAI/gemma-3-27b-it-FP8-dynamic | CO | 0.0501 | 0.0298 | 0.0294 | 0.0411 | −0.0255 | −0.0008 | 0.0125 | 0.0429 | root majority, 14 | 13 |
| RedHatAI/gemma-3-27b-it-FP8-dynamic | DA | 0.0205 | 0.0249 | 0.0343 | 0.0681 | 0.0041 | 0.0041 | 0.0076 | 0.0158 | full count, 21 | 15 |

Notes: Computed from the same logs as Table S5c, without a model call. $\Delta_X$ is the risk in arm R minus that in arm X (n/a in arm R). Counts are answers within 0.001 of a rule's output, out of 2,976 valid answers per arm, for the rules of Table S3f (results/open_weight_all_arms.json).

## S5.4 Condition A: withheld generating parameters

Conditions A and B measure how far raw baselines move when the prompt configuration and the run period change together (Section 7.3 gives the readings). They are exploratory additions made after the formal results were seen (deviations D4, D5 and D7; Section S6.2.3), with descriptive intervals. Both use a fixed subset of the main panel in arms R and G: the first worlds of each layer × cell in panel order, chosen before their first request. The subset has 998 worlds, with 252, 244, 256 and 246 at $\delta = 0, 0.3, 0.6$ and 1. The main panel on the subset, the two conditions and the open-weight configurations share one set of $B = 10{,}000$ world resamples, so paired differences are taken resample by resample. Each contrast with the main run is one joint difference in prompt and run period, to which condition A adds a change of channel for Opus and Haiku (Vertex AI, deviation D7; Section S5.5).

**Condition A** (stored as E-A in the study files) withholds the numerical values of a, e and δ and replaces items 2, 3 and 5 of the generating description with:

```
2. There are two sources. Each source has its own reading of the state. Given the state, the two
readings are independent, and each equals the state with a probability that is not given.
3. Each line below belongs to one source. An original report states its source's reading, flipped
with a probability that is not given; flips are independent across original reports given the
readings.
5. Taken alone, each original report is correct with probability r = 0.65. Original reports from
the same source can be correlated in whether they are correct. The strength of this correlation
is one of four ordered categories: low, moderate, high, very high. Here it is moderate.
```

**Table S5e.** Condition A, per configuration. Paired entries are at δ = 0.3 and 0.6.

| Configuration | Condition | D | $T_G$ | w3 | w2 | w3(0.3) − w3(0.6) |
|---|---|---|---|---|---|---|
| claude-opus-5-5 | A | 0.432 [0.340, 0.521] | 0.0002 [−0.0003, 0.0006] | 1.857, 1.358 | 1.005, 0.967 | 0.499 [0.434, 0.561] |
| gpt-6.1-sol | A | 0.516 [0.458, 0.570] | 2.9e−05 [−0.0001, 0.0002] | 1.714, 1.234 | 0.986, 1.011 | 0.479 [0.450, 0.509] |
| claude-haiku-5-5 | A | 0.231 [−0.055, 0.518] | 0.0074 [0.0018, 0.0132] | 2.279, 2.204 | 0.576, 0.658 | 0.075 [−0.052, 0.205] |
| gpt-6-luna | A | 0.003 [−0.030, 0.038] | 0.0026 [−0.0043, 0.0100] | 0.303, 0.273 | 0.126, 0.125 | 0.029 [−0.070, 0.133] |

Notes: The exact D is 0.428, and the exact w3 and w2 are those of Table S1b (w3 = 1.702 and 1.274 at δ = 0.3 and 0.6). Other rows, with v, Λ and their Fieller-type sets, are in results/explore_analysis.json.

**Table S5f.** Valid answers within 0.001 of a rule, arm R.

| Configuration | Condition | Arm | Valid | Failed | Exact posterior | First child | Prior |
|---|---|---|---|---|---|---|---|
| claude-opus-5-5 | main | R | 998 | 0 | 998 | 449 | 276 |
| gpt-6.1-sol | main | R | 998 | 0 | 998 | 449 | 276 |
| claude-haiku-5-5 | main | R | 998 | 0 | 375 | 253 | 206 |
| gpt-6-luna | main | R | 997 | 1 | 198 | 251 | 541 |
| gpt-6.1-sol | B | R | 997 | 1 | 996 | 449 | 276 |
| claude-haiku-5-5 | B | R | 998 | 0 | 75 | 73 | 90 |
| claude-opus-5-5 | A | R | 998 | 0 | 319 | 252 | 276 |
| gpt-6.1-sol | A | R | 998 | 0 | 435 | 300 | 276 |

Notes: For condition A, the exact posterior is the external full-information reference. The rows are arm R for the configurations cited in Section 7.3 and the failed items below. The other configurations and the arm G rows are in results/explore_analysis.json. The failed items are formal_main-r65-d3-p75-L1-0012 (gpt-6-luna, main, arm R, refusal; Table S2a) and formal_main-r65-d0-p75-L0-0037 (gpt-6.1-sol, B, arm R, unparsed). All open-weight answers on the subset are valid.

## S5.5 Condition B and channel comparisons

**Condition B** (stored as E-B in the study files) is also answered by the open-weight configurations, with the arm sentences unchanged. It replaces the answer instruction with:

```
Answer with only a JSON object of the form {"p_state_1": p}, where p is your probability (between
0 and 1) that the state is 1. Do not show your reasoning or write any other text.
```

From the main run to condition B, on the same worlds and arms, mean output tokens change from 442.9 to 284.0 for Opus (median visible text from 557 to 21 characters), from 159.2 to 156.0 for Sol, from 188.7 to 15.1 for Haiku and from 17.9 to 16.9 for Luna. Luna's D1 refusal in the main run returned no output and is excluded from its mean.

**Table S5g.** Four-configuration means.

| Condition | $T_G$ | $D_R$ | w3(0.3) − w3(0.6) | Λ | v gates in arm R |
|---|---|---|---|---|---|
| Main panel on the subset, four configurations | −0.0009 [−0.0053, 0.0038] | 0.261 [0.132, 0.389] | 0.340 [0.242, 0.441] | n/a | yes, yes, yes, no |
| Condition B, four configurations | −0.0003 [−0.0056, 0.0050] | 0.292 [0.153, 0.432] | 0.262 [0.156, 0.366] | 0.075 [0.023, 0.123] | yes, yes, yes, yes |
| B minus main, four configurations | 0.0007 [−0.0057, 0.0071] | 0.031 [−0.152, 0.216] | −0.078 [−0.232, 0.075] | n/a | n/a |

Notes: n/a for Λ: a v gate fails on the subset, or the row is a difference. Gate entries follow claude-opus-5-5, gpt-6.1-sol, claude-haiku-5-5 and gpt-6-luna. The rows for condition A and A minus main are in results/explore_analysis.json.

**Table S5h.** Paired differences per configuration.

| Configuration | Difference | D | $T_G$ | w3, δ = 0.3 | w3, δ = 0.6 |
|---|---|---|---|---|---|
| claude-haiku-5-5 | B minus main | −0.278 [−0.909, 0.348] | 0.0152 [0.0008, 0.0291] | −0.067 [−0.377, 0.249] | 0.189 [−0.016, 0.395] |
| gpt-6-luna | B minus main | 0.400 [0.019, 0.799] | −0.0113 [−0.0331, 0.0100] | 0.255 [−0.049, 0.575] | 0.311 [−0.015, 0.641] |

Notes: B minus main is the joint difference in answer sentence and run period of Section S5.4. The rows for the other configurations and for A minus main are in results/explore_analysis.json.

**Run record.** Conditions A and B and the channel comparisons cost USD 42.48 against the joint cap of USD 50. The itemised costs and the output lengths of conditions A and B are computed from the run logs (Section S6.4). The main panel and condition B ran through the vendor Batch interfaces, and Sol and Luna ran condition A through the OpenAI Batch interface. Condition B was complete on 31 October 2026 UTC and condition A on 1 November 2026 UTC, and all 1,996 requests of each configuration in condition A returned valid answers (Table S5f).

**Channel comparison** (deviation D7; 200 repeated condition B prompts per configuration).

**Table S5i.** Channel comparison on condition B prompts, Message Batches API against Vertex AI online requests.

| Configuration | Items | Valid on both | Within 0.001 | Mean absolute difference | Largest absolute difference |
|---|---|---|---|---|---|
| claude-opus-5-5 | 200 | 200 | 198 | 0.00157 | 0.1619 |
| claude-haiku-5-5 | 200 | 200 | 102 | 0.04370 | 0.4500 |

Notes: Exploratory; deviation D7 (Table S6b).

## S6. Registration, deviations and reproducibility records

This section dates the registration, records the deviations D1–D8 and the code corrections C1–C6 (summary in Section 4.4 of the article), and fixes the configurations, execution settings and archive entry points. The registered analysis definitions are unchanged.

## S6.1 Development and registration chronology

**Table S6a.** Chronology of development, registration and formal runs (2026).

| Date | Event |
|---|---|
| 23 October | Simulated studies per calibration scenario raised from 10,000 to 20,000, after an interim summary under plan v3.4 |
| 25–26 October (UTC) | Admission check, 24 items per configuration: 24 of 24 valid answers and a single model field from each configuration |
| 26 October | Pilot: 60 naturally sampled worlds per δ under all four arms (960 requests per configuration) and 48 wording-control requests per configuration; admission and pilot cost USD 13.80. Plan v3.5 decided before any formal data existed |
| 27 October | γ reference zones proposed, after the discrimination simulations of Section S3.5 were seen |
| 28 October | After the v3.5 coverage results were seen, scenarios A09, C04 and C05, which did not meet the target (39 of 42 gating scenarios passed), were reclassified and the calibration-support screen was added. γ zones adopted, model registry frozen, study preregistered on OSF (registration dzgj2; access is restricted); formal requests began once the registration was complete |
| 29 October | Corresponding author's decision on D1 |
| 31 October | Closed-weight matrix and open-weight replication complete; the analysis driver first ran on formal answers |

The English registration form summarises the design, and the attached plan, analysis specification and execution rules govern where they differ. The formal analysis driver ran on simulated data with known answers before registration.

**Development.** Development data come from seed blocks and worlds disjoint from the formal ones, were collected under plan v3.4 and are kept apart from every confirmatory and secondary estimate. The pilot ran without transport failures. Its stop triggers, fixed before it was run, were met in 23 instances and stopped plan v3.4 as the plan prescribes. The single precision trigger traces to two answers of exactly 0 from one configuration. At $\varepsilon = 10^{-6}$ their logit of −13.8 dominated the predicted variance of $D_R$.

**Plan v3.5.** Plan v3.5 sets confirmatory truncation at $\varepsilon = 0.001$ and keeps the v3.4 scale, $\varepsilon = 10^{-6}$, as a prespecified sensitivity analysis (Section S3.2). The materials, the four configurations, the confirmatory family, the number of worlds and the registered reporting bounds carry over from v3.4. The v3.5 triggers were set and assessed on the same pilot data, and none fired. The γ reference zones have half-width 0.05 because truncation and four-decimal rounding shift the γ of an exact receiver by about 0.0002, while its standard error is about 0.0001.

## S6.2 Deviations D1–D8

Eight deviations from the registered rules occurred after registration. Sections S6.2.1–S6.2.3 give the details behind the table.

**Table S6b.** Deviations D1–D8 from the registered rules.

| ID | Registered rule | What was done | Reason | Reading scope |
|---|---|---|---|---|
| D1 | An OpenAI Batch item rejected with a client error stops its block; a refusal is a semantic failure | Input-side rejections of gpt-6-luna requests are recorded at once as refusals, and the block continues | The vendor's content filter acts on the input rather than the output (corresponding author, 29 October 2026) | Nine refusals (Table S2a); excluded from the complete-case version of the confirmatory family |
| D2 | Open-weight replication on the registered host, a laboratory A6000 machine | One H100 on Google Cloud in Tokyo (a3-highgpu-1g, Flex-start), vLLM 0.32.0, registered model snapshots | The registered host had been offline since 28 October 2026 | Descriptive replication |
| D3 | One bounded format correction is allowed, with all models admitted again under it | Template P0-draft-3 adds the sentence "Do not show your reasoning or write any other text." for the open-weight configurations | Truncation at the registered visible cap of 256 tokens at admission | Open-weight configurations; the confirmatory panels stay under P0-draft-2 |
| D4 | None | Condition A: dependence given as an ordered verbal level, with a, e and δ withheld | Probes raw weighting when the generating parameters are withheld | Exploratory (Section S5.4) |

| ID | Registered rule | What was done | Reason | Reading scope |
|---|---|---|---|---|
| D5 | None | Condition B: the closed-weight configurations receive the answer sentence of P0-draft-3 | Tests the sensitivity of raw baselines to the answer sentence | Exploratory; contrasts with the main run span answer sentence and run period (Section S5.5) |
| D6 | None | Scenario X04, a coverage scenario for the formal screen composition, built after the formal results were seen | The confirmatory reading of $D_R$ extrapolates the pre-registration calibration to the observed composition (three varying configurations, one nearly constant) | Coverage and power under nonzero true contrasts (Section S3.3; Table S3j) |
| D7 | Vendor Batch interfaces, as for the formal matrix | Opus and Haiku ran condition A through Vertex AI as online requests, with 200 repeated condition B prompts compared (Table S5i) | The first condition A batches of claude-opus-5-5 and claude-haiku-5-5 were rejected because of an insufficient account balance | Exploratory; contrasts with the main run combine channel and run period (Sections S5.4 and S5.5) |
| D8 | None | Table S5d: point estimates for the open-weight configurations in all four main-panel arms, computed from the existing formal logs without a model request | Post hoc descriptive reproduction of the gain pattern (Section 6.6) | Descriptive (Section S5.3) |

### *S6.2.1 D1: input-side content filter*

In the formal run, OpenAI rejected four gpt-6-luna requests at input with HTTP 400 and error code invalid_prompt (the prompt was judged a possible violation of the usage policy). Two were in the main panel (arms R and G), one in arm G of the $r = 0.70$ panel and one, a copy-slot request, in arm G of the wording-control panel. The three affected blocks stopped as registered. From the decision of 29 October 2026, such an item is recorded at once as a refusal (a semantic failure with $q = \pi$ and a flag, with no replay and no cost), and its block continues. The four earlier rejections counted as no attempt, as the registered rules provide. Each was resent once and returned a valid answer.

### *S6.2.2 D2 and D3: open-weight hardware and answer sentence*

**D2.** Requests carry no sampling parameters and no seed, so sampling follows the generation configuration that each server applied. The Qwen server log records its effective defaults. For Gemma the record is its stored generation configuration, and the effective sampling scalars beyond it were not recorded. Item-level answers are archived as returned by this run on these snapshots, hardware and server version, and every analysis reruns from them.

**D3.** At admission on the Google Cloud machine, Qwen was truncated at the visible cap on 2 of 21 answers and Gemma on all 5, and their blocks stopped. The four closed-weight configurations had already completed their formal panels under template P0-draft-2 and kept them, so the re-admission that the registered rules require applied to the open-weight configurations. Both passed a new admission with 24 of 24 valid answers and input token counts equal to the tokenizer check on 24 of 24 items. The replication ran on 31 October 2026 and returned 14,400 valid answers per configuration.

### *S6.2.3 D4–D8: post-registration additions*

D4–D6 and D8 were added after the formal results were seen, and D7 concerns the execution of condition A. All five sit outside the confirmatory family and leave the registered decisions unchanged (Table S6b).

## S6.3 Corrections C1–C6

Six corrections to the registered code preserve the registered analysis definitions (Table S6c).

**Table S6c.** Corrections C1–C6 to the registered code.

| ID | Correction and scope |
|---|---|
| C1 | The driver reports each open-weight configuration by its own state (results, incomplete or not run), so the confirmatory family rests on the four closed-weight configurations (shared with C5) |
| C2 | A final outcome classification per panel and configuration is added, with the D1 items listed |
| C3 | The driver tabulates the most frequently matched program rule for arms R, G, CO and DA, as the specification requires; the calibration-support screen uses arms R and G, as registered |
| C4 | A back end for the registered vLLM runs is added, written to the registered rules |
| C5 | Open-weight state reporting, as under C1 |
| C6 | Differences are compared at 12 decimals, so the seven gpt-6-luna answers that differ from a rule's output by exactly 0.001 count as matches in the rule-match table (Table S3f); the effect is on that descriptive table, and the agreement definition of Section S2.3 uses the same comparison |

C1, C2 and C4 were checked by production-format runs on simulated responses.

## S6.4 Configurations, execution and archive contents

**Configurations.** The six configurations are deployed units of model and vendor setting, fixed across arms and wording frames. A shared setting name such as low effort names a vendor parameter. Table 2 gives reasoning settings, templates and output caps, and Table S6d gives prices and snapshots.

**Table S6d.** The four closed-weight configurations: prices and snapshots.

| | Standard price (USD per million tokens) | | Batch price (USD per million tokens) | | |
|---|---|---|---|---|---|
| **Model ID** | **Input** | **Output** | **Input** | **Output** | **Snapshot and retirement** |
| claude-opus-5-5 | 4.00 | 20.00 | 2.00 | 10.00 | Pinned snapshot; retirement not before 22 September 2027 |
| gpt-6.1-sol | 2.00 | 10.00 | 1.00 | 5.00 | Default snapshot, the only one listed; no deprecation date listed |
| claude-haiku-5-5 | 0.10 | 0.50 | 0.05 | 0.25 | Pinned snapshot; retirement not before 7 October 2027 |
| gpt-6-luna | 0.10 | 0.50 | 0.05 | 0.25 | Default snapshot, the only one listed; no deprecation date listed |

Notes: Registry frozen on 28 October 2026. Snapshot and retirement information is from vendor pages read on 28 October 2026. Reasoning settings are in Table 2.

**Execution.** Formal requests go through the Anthropic Message Batches API and the OpenAI Batch API (/v1/responses endpoint, 24-hour window) at Batch prices, in blocks of configuration × panel. Requests within a block are submitted in a seeded random order registered in advance, so that arms, dependence levels and priors are interleaved in time. A different returned model field, a failure rate above 5%, an exhausted panel replay reserve, a request rejected before processing or a retired endpoint stops the block. A hard cost cap of USD 130 is enforced by reserving the worst-case cost of each batch before submission.

**Reproducibility manifest.** The file submission/configuration_reproducibility_manifest.json documents the six configurations from the existing registry, request and response logs and serving records, without new endpoint queries. It also lists three categories of unavailable records: an independent immutable vendor weight or build identifier for the closed-weight configurations, the served tokenizer revision and class, and the effective Gemma sampling scalars beyond its stored generation configuration (Section S6.2.2). The manifest's replay claim is the analysis of the archived responses, which reruns without new generation. A stochastic regeneration is a new run of the configurations the manifest documents.

**Archive entry points.** Tables that show selected arms point to the review archive for complete values: results/formal_analysis.json (cpu/formal_analysis.py) for the formal analysis, results/explore_analysis.json (cpu/explore_analysis.py) for conditions A and B, and results/open_weight_all_arms.json (cpu/open_weight_all_arms.py) for the open-weight all-arm runs. The cost itemisation and output lengths of conditions A and B are in results/explore_run_record.json (cpu/explore_run_record.py).